\documentclass{article}

\usepackage[final]{neurips_2025}

\usepackage[utf8]{inputenc} 
\usepackage[T1]{fontenc}    
\usepackage{hyperref}       
\usepackage{url}            
\usepackage{booktabs}       
\usepackage{amsfonts}       
\usepackage{nicefrac}       
\usepackage{microtype}      
\usepackage{xcolor}         

\usepackage{float}
\usepackage{graphicx}
\usepackage{tablefootnote}

\usepackage{amsmath}
\usepackage{amssymb}
\usepackage{mathtools}
\usepackage{amsthm}

\usepackage{authblk}

\DeclareMathAlphabet{\mathbbold}{U}{bbold}{m}{n}
\newcommand{\bigbar}[1]{\overline{#1}}

\setcitestyle{authoryear,round,citesep={;},aysep={,},yysep={;}}

\title{REVE: A Foundation Model for EEG \\ Adapting to Any Setup with \\ Large-Scale Pretraining on 25,000 Subjects}

\author{
Yassine El Ouahidi$^{1}$\thanks{Corresponding authors: \texttt{yassine.elouahidi@mistral.ai}, \texttt{giulia.lioi@imt-atlantique.fr}},
Jonathan Lys$^{1}$,
Philipp Thölke$^{2}$,
Nicolas Farrugia$^{1}$,\break
Bastien Pasdeloup$^{1}$,
Vincent Gripon$^{1}$,
Karim Jerbi$^{2, 3, 4}$,
Giulia Lioi$^{1*}$\\
\small $^{1}$ IMT Atlantique, Lab-STICC, UMR CNRS 6285, F-29238 Brest, France\\
\small $^{2}$ Psychology Department, Université de Montréal, Montreal, QC, Canada\\
\small $^{3}$ Mila (Quebec AI research institute), Montreal, QC, Canada\\
\small $^{4}$ UNIQUE (Quebec Neuro-AI research center), QC, Canada
}

\begin{document}





\maketitle

\begin{abstract}
Foundation models have transformed AI by reducing reliance on task-specific data through large-scale pretraining. While successful in language and vision, their adoption in EEG has lagged due to the heterogeneity of public datasets, which are collected under varying protocols, devices, and electrode configurations. Existing EEG foundation models struggle to generalize across these variations, often restricting pretraining to a single setup, resulting in suboptimal performance, in particular under linear probing.
We present REVE (Representation for EEG with Versatile Embeddings), a pretrained model explicitly designed to generalize across diverse EEG signals. REVE introduces a novel 4D positional encoding scheme that enables it to process signals of arbitrary length and electrode arrangement. Using a masked autoencoding objective, we pretrain REVE on over 60,000 hours of EEG data from 92 datasets spanning 25,000 subjects, representing the largest EEG pretraining effort to date.
REVE achieves state-of-the-art results on 10 downstream EEG tasks, including motor imagery classification, seizure detection, sleep staging, cognitive load estimation, and emotion recognition. With little to no fine-tuning, it demonstrates strong generalization, and nuanced spatio-temporal modeling. We release code, pretrained weights, and tutorials\footnote{ Project page: \url{https://brain-bzh.github.io/reve/}} to support standardized EEG research and accelerate progress in clinical neuroscience.
\end{abstract}

\section{Introduction} 
\label{sec:introduction}

Electroencephalography (EEG) is a non-invasive technique widely used to study brain activity, with applications spanning brain-computer interfaces (BCIs), clinical diagnostics, and neuroscience research. 
Despite its potential, the adoption of EEG-based technologies remains limited~\citep{lotte2018review}. A key challenge is developing models that generalize effectively to new subjects. EEG data varies widely in electrode configurations, recording conditions, and subject-specific factors, complicating model transferability. 
This heterogeneity has led to a fragmented ecosystem of datasets and task-specific models, many of which struggle to generalize across settings.

Foundation models have transformed natural language processing~\citep{achiam2023gpt4, dubey2024llama3, warner2024modernbert} and computer vision~\citep{radford2021learning, caron2021emerging, kirillov2023segment} by leveraging large-scale pretraining to enable transfer with minimal supervision. Their ability to produce general-purpose representations has sparked growing interest in building similar models for EEG~\citep{yang2024biot, wang2024cbramod, jiang2024labram, cui2024neuro, yuan2024brant2, wang2024eegpt}. Yet, EEG poses unique challenges including data heterogeneity, low signal-to-noise ratio, and the lack of standardized positional encoding to accommodate varying electrode configurations.

Recent EEG foundation models such as BIOT~\citep{yang2024biot}, Labram~\citep{jiang2024labram}, CBraMod~\citep{wang2024cbramod}, and NeuroGPT~\citep{cui2024neuro} adopt self-supervised learning (SSL) techniques for pretraining. While promising, many of these models rely solely on the TUH database~\citep{obeid2016temple} which uses a fixed 19 or 21-channel montage. As a result, they often fail to generalize to datasets with different electrode layouts or recording setups. Furthermore, existing positional encoding schemes, whether absolute~\citep{yang2024biot, jiang2024labram} or convolutional~\citep{wang2024cbramod}, lack the flexibility to accommodate spatial diversity, often necessitating full fine-tuning for transfer.



To address the limitations in current EEG foundation models, we consider three core contributions that enable scalable, generalizable representation learning across diverse, large-scale EEG datasets. 

First, we propose a novel 4D positional encoding scheme that enables flexible modeling of EEG signals with varying temporal lengths and electrode configurations. Unlike existing absolute or convolutional encodings, our formulation naturally supports spatial and temporal variability, eliminating the need for fixed montages or fine-tuning of positional priors.

Thanks to this flexible positional encoding method, we are able to train with a wider range of EEG configurations, allowing to scale to larger and more heterogeneous datasets. To this end, we curate the largest and most diverse EEG corpus to date, comprising over 60,000 hours of data from 92 datasets and 25,000 subjects. This diverse collection spans clinical, BCI, and research domains, providing the scale and diversity necessary for robust pretraining.

Combining architectural flexibility with large-scale data results in REVE (Representation for EEG with Versatile Embeddings), a spatio-temporal transformer model trained with a modified masked autoencoder (MAE)~\citep{he2022masked} objective that promotes learning better representations in the model. REVE learns general-purpose EEG representations that transfer effectively across a wide range of downstream tasks.



REVE achieves state-of-the-art performance across numerous benchmarks, including BCI and clinical datasets, outperforming prior EEG foundation models. Our scaling studies further show improved generalization with larger model sizes, reinforcing the benefits of large-scale pretraining. To support adoption, we release open-source code, pretrained models of multiple sizes, and detailed tutorials for applying REVE to various EEG tasks. By addressing the unique challenges of EEG with scalable architectures and flexible spatial encoding, REVE establishes a unified foundation for EEG analysis and paves the way for new advances in neuroscience and clinical applications.

\section{Methods}
\label{sec:method}

\begin{figure}[h]
\centering
\includegraphics[width=.92\linewidth]{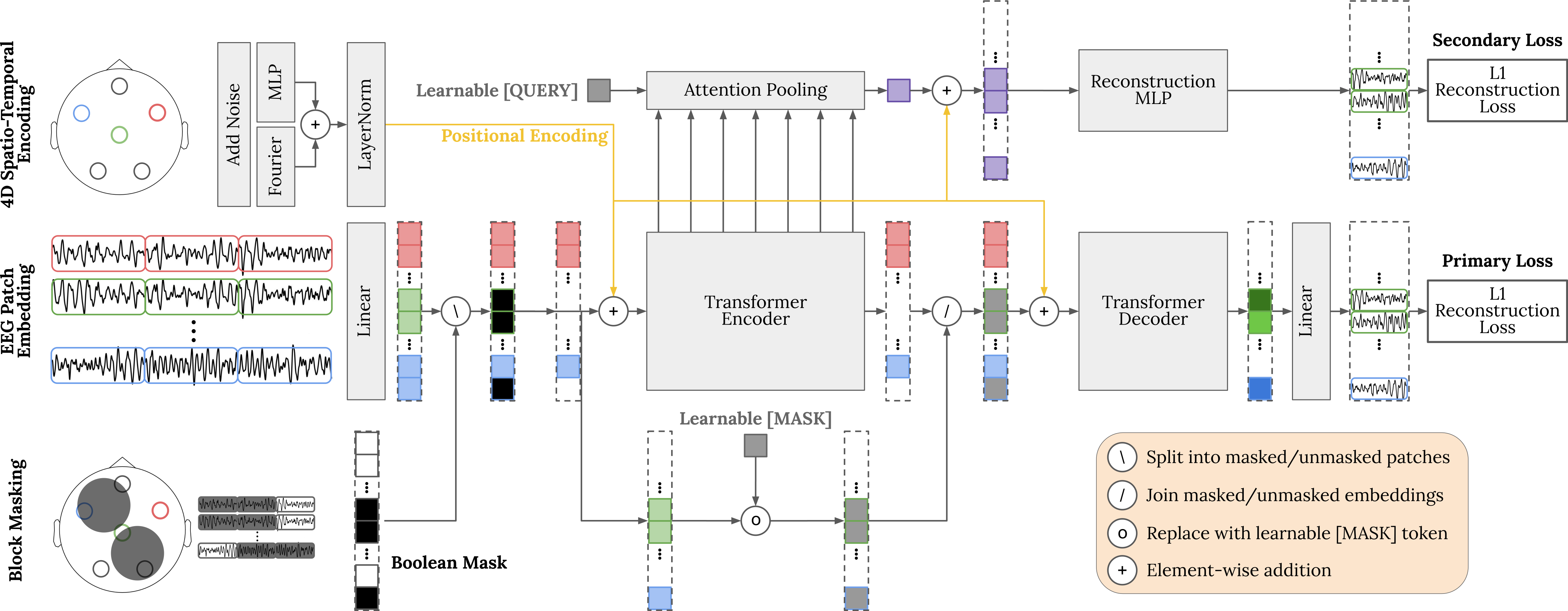}
\caption{Overview of the \textbf{REVE} pretraining framework. The model processes multi-channel EEG data through a linear \textbf{Patch Embedding} where signals are divided into overlapping temporal patches for each channel and embedded with a linear layer. \textbf{4D Spatio-Temporal Position Encoding} combines spatial coordinates of electrodes with temporal patch indices, augmented with noise for robust generalization. A \textbf{Block Masking Strategy} masks contiguous regions across spatial and temporal dimensions to simulate realistic disruptions. The transformer encoder processes unmasked embeddings. Updated embeddings are joined with learnable placeholders for the masked tokens, from which raw EEG is reconstructed using the decoder. The \textbf{Primary Task} predicts raw EEG signals directly, while the \textbf{Secondary Task} trains a single global token via attention pooling to summarize the input. Both tasks minimize an $\text{L}_1$    reconstruction loss.}

\label{fig:archi}
\end{figure}
We pretrain our encoder using a masked autoencoder objective. The REVE encoder consists of a patch embedding module, a 4D position encoding module, and a transformer backbone. During pretraining, we apply spatio-temporal contiguous masking to the patch embeddings and jointly train the encoder and decoder to reconstruct the missing segments of EEG, enabling the encoder to learn robust feature representations. Subsequent hyperparameter values are listed in Table~\ref{tab:summary} in the Appendix.

\subsection{EEG Representation and Block Masking strategy}
\label{subsec:patching}
\label{subsec:mask-strat}


We represent multi-channel EEG data as $\mathbf{X} \in \mathbb{R}^{C \times T}$, where $C$ is the number of electrodes and $T$ the number of time samples, electrode positions are given by $\mathbf{P} \in \mathbb{R}^{C \times 3}$, corresponding to their 3D coordinates. To process the data, we segment each channel into patches of size $w$ with overlap $o$, following BIOT~\citep{yang2024biot}. This yields $p = \left\lceil \frac{T - w}{w - o} \right\rceil + \mathbbold{1} \left[ (T - w) \bmod (w - o) \neq 0 \right]$ non-overlapping patches (discarding any incomplete ones), and reshapes $\mathbf{X}$ into $\mathbf{Xp} \in \mathbb{R}^{C \times p \times w}$. Each patch is linearly embedded, resulting in $\mathbf{E} \in \mathbb{R}^{C \times p \times D_E}$, where $D_E$ is the embedding dimension.


To enhance learning during pretraining, we apply a joint spatio-temporal block masking strategy that masks structured regions across both spatial and temporal dimensions. Random masking, proposed for EEG by~\citet{chien2022maeeg}, was later improved through spatial masking~\citep{mohammadi2024eeg2rep, guetschel2024s}.  In this work, we extend the masking strategy to the temporal domain. This builds on insights from image modeling, where structured masking outperforms random masking~\citep{xie2022simmim}, a trend also supported by our ablation results (Table~\ref{tab:rmask_results}, Appendix). As neighboring segments of EEG, in both spatial and temporal domain, are typically similar, naive random masking could leave redundant information exposed, reducing the difficulty of reconstruction. In contrast, block masking better disrupts these patterns, encouraging more effective learning.


Our block masking strategy is governed by the following parameters: The masking Ratio $M_r$ controls the overall proportion of masked tokens. The spatial Masking Radius $R_s$ and Temporal Masking Radius $R_t$ respectively define the spatial extent (around a selected channel) and the time window (around a selected token) to be masked. Similarly, the Dropout Ratio $D_r$ sets the fraction of masked tokens for which the entire time series of the corresponding channel is dropped, while the dropout Radius $R_d$ determines the spatial neighborhood affected by dropout. For tokens not dropped, temporal masking is applied within radius $R_t$. This process yields a binary mask $\mathbf{B} \in \mathbb{R}^{C \times p}$, containing $N_{\text{m}} = \lfloor (1 - M_r) \cdot C \cdot p \rfloor$ masked entries (zeros) and $N_{\bar{\text{m}}} = C \cdot p - N_{\text{m}}$ unmasked entries (ones).

\subsection{4D Position Encoding Strategy}
\label{subsec:fourier}

Unlike prior works that rely on learned embedding tables for spatial encoding vectors~\citep{jiang2024labram, wang2024cbramod}, we directly generate position encodings from the spatio-temporal coordinates of the tokens, allowing the processing of signals of any length or EEG layout and enabling better generalization to unseen setups. More specifically, our method uses a transformation applicable to each position, utilizing the actual 3D coordinates and timestep of each EEG patch, enabling the model to handle arbitrary electrode configurations and sequence lengths without relying on learned embeddings.

\paragraph{4D Positional Encoding and Spatial Augmentation.}
We start with the spatial positions of the EEG electrodes $\mathbf{P} \in \mathbb{R}^{C \times 3}$, where each row of \(\mathbf{P}\) contains the \((x, y, z)\) coordinates of a channel, to which we add Gaussian noise with standard deviation \(\sigma_{\text{noise}}\). This improves generalization to diverse electrode positions and ensures robustness to variability in head size or electrode placement. We extend $\mathbf{P}$ with a temporal component, resulting in $\mathbf{P}_{\text{ext}} \in \mathbb{R}^{C \times p \times 4}$,
where $p$ is the number of patches obtained from segmenting EEG signal, as defined in Section~\ref{subsec:patching}. The temporal dimension is represented as discrete values from \(1\) to \(p\), scaled by a factor $s_{\text{t}}$ to ensure a scale similar to the spatial dimensions.

\paragraph{4D Fourier-Based Position Encoding.}

Building on the 2D approach proposed by~\citet{defossez2023decoding}, we extend the Fourier positional encoding method to 4D in our encoding strategy, as follows. Each positional component $(x,y,z,t)$ of $\mathbf{P}_{\text{ext}}$ is projected into a multi-frequency space, using $n_{\text{freq}}$ frequencies per dimension. The frequency assignment follows a Cartesian product structure, \emph{i.e.}, all combinations of frequencies across the four dimensions contribute to the encoding, resulting in a flattened vector of dimension $n_{\text{freq}}^4$.
A hierarchical periodicity emerges: the period of $x$ is $n_{\text{freq}}^1$, of $y$ is $n_{\text{freq}}^2$, of $z$ is $n_{\text{freq}}^3$, and of $t$ is $n_{\text{freq}}^4$. Then, applying sine and cosine transformations doubles the embedding size, producing a positional vector of dimension $2\cdot n_{\text{freq}}^4$. We ensure that the  embedding dimension matches the hidden size required by the 4DPE module, with $n_{\text{freq}} \in \{3, 4, 5\}$ resulting in the final embedding $\mathbf{F}_{\text{pe}} \in \mathbb{R}^{C \times p \times D_E}$. The 4D encoding adds minimal compute overhead, with sinusoidal computations and a small linear layer. Computational cost scales linearly with the number of input tokens (channels $\times$ temporal patches) and is negligible relative to the transformer backbone.



\paragraph{Final Adjusted Position Encoding.}
To complement the fixed Fourier features, we also process $\mathbf{P}_{\text{ext}}$ through a linear layer followed by GELU~\citep{hendrycks2016gaussian} and LayerNorm~\citep{lei2016layernorm}, producing a learnable representation $\mathbf{F}_{\text{lin}} \in \mathbb{R}^{C \times p \times D_E}$. This component adapts the positional encoding to the specific dataset and task, and can compensate for any truncation in the Fourier basis. The final positional encoding is given by $\mathbf{P}_{\text{enc}} = \text{LayerNorm}(\mathbf{F}_{\text{pe}} + \mathbf{F}_{\text{lin}})$, combining the structured inductive bias of Fourier features with the flexibility of learned adaptation. This vector is added to the non-masked patch embeddings before being passed to the transformer encoder similarly to MAE~\citep{he2022masked}, and is consistent with standard absolute positional encoding practices~\citep{vaswani2017attention}. The ablation study in Table~\ref{tab:ablation_pe} confirms that this method outperforms both fixed learnable and purely MLP-based positional encoding schemes.

\subsection{Transformer}
\label{subsec:transfo}
Our model extends the standard Transformer architecture~\citep{vaswani2017attention} with enhancements that improve efficiency and stability.
We use RMSNorm~\citep{zhang2019root} in lieu of LayerNorm as a \textbf{normalization layer} for better training stability, and choose GEGLU~\citep{shazeer2020glu} as the \textbf{activation} function in the feed-forward network (FFN) layers as it outperforms standard GELU through more expressive gating mechanisms~\citep{geiping2023cramming}. This choice is further supported by the ablation results in Table~\ref{tab:ablation_activation}.
Our \textbf{FFN layers} follow a two-layer structure with an expansion ratio of \( \frac{8}{3} \), consistent with designs from LLaMA~\citep{touvron2023llama1}, Qwen~\citep{bai2023qwen} or Mistral~\citep{jiang2023mistral}.
Following~\citet{Dayma_DALLE_Mini_2021}, we \textbf{remove bias terms} from all linear layers except the final decoder layer. This reallocates the parameter budget to linear transformations, improving efficiency. We use \textbf{Flash Attention v2}~\citep{ dao2023flashattention2} for memory and computational efficiency in the attention as it reduces the softmax overhead and ensures scalability to long sequences, while maintaining the core transformer formulation. 

\subsection{Masked EEG Reconstruction Methodology}


During pretraining, our model reconstructs EEG signal of masked patches using information from the visible, unmasked patches. The overall pretraining framework is illustrated in Figure~\ref{fig:archi}. 

Let $\mathbf{P}_{\text{m}} \in \mathbb{R}^{N_{\text{m}}\times w}$, and $\mathbf{P}_{\bigbar{\text{m}}} \in \mathbb{R}^{N_{\bigbar{\text{m}}} \times w}$ denote the masked and visible patches, respectively, with $N_{\text{m}}$ and $N_{\bigbar{\text{m}}}$ as defined in Section \ref{subsec:mask-strat}. The associated patch embeddings are denoted as $\mathbf{E_m}$ for masked patches and  $\mathbf{E_{\bigbar{m}}}$ for visible patches.


We adopt the MAE structure from~\citet{he2022masked}, with a larger encoder and a lighter decoder each following the architecture described in Section~\ref{subsec:transfo}. Only the embeddings of the visible patches $\mathbf{E}_{\bigbar{\text{m}}}$, enriched with their positional encodings are passed through the encoder, to produce latent representations $\mathbf{F}_{\bigbar{\text{m}}}$. 
Masked patches are represented using a learned embedding, repeated $N_{\text{m}}$ times and also augmented with positional encodings.
Before entering the decoder, positional encodings are re-added to both visible and masked latent patches. Together, they form the decoder input from which the raw EEG signal of the masked patches is reconstructed.

Unlike the original MAE, which uses a separate set of fixed positional encodings for the decoder, we reuse the same encoding for both the encoder and decoder. This design ensures flexibility for processing EEG signals with varying temporal lengths and electrode configurations.

The output of the decoder transformer, is passed through a linear projection layer that maps latent patches back into the signal space, reconstructing the raw EEG signal of the masked patches. Reconstructed patches minimize the $\text{L}_1$ loss relative to the original raw EEG patches:
\begin{equation}
    \mathcal{L} = \frac{1}{|\mathbf{P}_{\text{m}}|} \sum_{i \in \mathbf{P}_{\text{m}}} \left\lVert \mathbf{\hat{P}}_{\text{m}}^{(i)} -\mathbf{P}_{\text{m}}^{(i)} \right\rVert_1
\end{equation}
where $\mathbf{\hat{P}}_{\text{m}}^{(i)}$ represents the reconstructed signal for patch $i$, and $\mathbf{P}_{\text{m}}^{(i)}$ is the original signal. We chose $\text{L}_1$ loss over $\text{L}_2$ due to the inherently noisy nature of EEG signals. While $\text{L}_2$ amplifies the influence of noise, $\text{L}_1$ loss offers greater robustness by reducing the impact of outliers.


In addition to the main reconstruction loss, we introduce a secondary task that reconstructs masked patches from a compact global representation. We apply attention pooling over the outputs of all Multi-Head Attention (MHA) layers in the encoder: the output tokens (after FFN) from each MHA block are concatenated and attended by a learned query token. This pooled token is then repeated, enriched with positional encodings, and passed through a 2-layer FFN to reconstruct the masked patches. As with the primary loss, we use $\text{L}_1$ loss for reconstruction. The total loss is a weighted sum: $\text{Loss} = \text{Primary Loss} + \lambda \cdot \text{Secondary Loss}$

This secondary loss encourages the encoder to distribute useful information across all layers, mitigating over-specialization in the final layer and yielding more generalizable representations.


The secondary loss mitigates a limitation of the MAE framework: the final encoder layer can overfit to the reconstruction task, especially with a shallow decoder~\citep{he2022masked}. By pooling features across all transformer layers~\citep{alkin2024mim}, the learned token captures a compact, global EEG representation, encouraging more balanced use of the encoder depth. This leads to stronger, more generalizable features for downstream tasks like linear probing, few-shot learning, and transfer without fine-tuning.

After the pretraining phase, the decoder is discarded, and only the encoder is used. In this case, no embeddings are masked, \emph{i.e.}, \( \mathbf{P}_{\text{m}} = \mathbf{E}_{\text{m}} =  \emptyset \). All patches are processed as usual by retaining their associated positional encoding.

To avoid confusion regarding terminology, we clarify that the terms “encoder” and “decoder” are used here in the context of masked auto-encoders (MAE), not in the autoregressive Transformer sense. All Transformer blocks in REVE are non-causal and operate within a standard encoder-style attention pattern; no autoregressive training is involved. The “decoder” refers solely to the lightweight reconstruction head used to recover masked EEG segments during self-supervised pretraining.
\section{Experiments}
\label{sec:exp_results}

\subsection{Pretraining}

This section outlines the data sources and preprocessing steps used for pretraining, followed by our strategy for scalable and effective representation learning across diverse datasets.

\subsubsection{Dataset Collection \& Preprocessing}



To enable large-scale pretraining, we assembled a massive and diverse collection of EEG recordings from open-source or request-accessible datasets. It comprises 19 TB of raw data, spanning 24,274 subjects, 150,833 unique sessions, and 61,415 hours of recordings drawn from 92 different sources, including OpenNeuro~\citep{markiewicz2021openneuro}, MOABB~\citep{Aristimunha_Mother_of_all_2023}, and TUH~\citep{obeid2016temple}. To our knowledge, this is the largest and most diverse EEG dataset assembled for training a foundation model. The most extensive prior effort, by~\citet{yuan2024brainwavebrainsignalfoundation}, comprised approximately 40,000 hours of recordings, but primarily relied on intracranial EEG (iEEG) rather than non-invasive EEG. A summary of the dataset composition and a full list of included sources are provided in Appendix~\ref{app:datasets}. While the majority of the data consists of clinical EEG recordings, we also include a substantial subset of cognitive and BCI-related data which, although smaller in proportion, tend to be cleaner and more diverse.
We also collected electrode positional information for each recording. When 3D coordinates were available, they were used directly; otherwise, positions were inferred from standard labels. Channels without identifiable names or positional data were excluded. The dataset spans a wide range of EEG systems and formats—including BrainVision, BioSemi, EDF, GDF, and EEGLAB, with most recordings adhering to the 10-5 system~\citep{oostenveld2001five}. In total, the dataset includes 396 unique electrode names.

Our preprocessing pipeline is designed to preserve signal diversity and prioritize robustness when scaling. We only removed recordings shorter than 10 seconds, and those used in downstream tasks. Remaining signals were resampled to 200 Hz, band-pass filtered (0.5–99.5 Hz), and converted to float32, resulting in a 6 TB dataset. To address amplitude variations across recordings, we applied Z-score normalization with statistics computed across the recording sessions to ensure robust statistics. After normalization, values exceeding 15 standard deviations were clipped, as in~\citet{defossez2023decoding}. Unlike CBraMod~\citep{wang2024cbramod}, which excluded signals above 100 $\mu\text{V}$, our approach retains them, resulting in about 60,000 hours of EEG, compared to 9,000 in CBraMod and 2,534 in LaBraM.

\subsubsection{Pretraining Strategy \& Scaling}

\subsection{Training and Scaling Strategy}

We present the training procedure used for pretraining the Small model and detail how it scales to larger architectures under constrained resources.
Our training framework builds upon recent advances in state-of-the-art NLP methodologies~\citep{warner2024modernbert}. We use the StableAdamW~\citep{wortsman2023stable} optimizer, designed for low precision frameworks and improved stability, thanks to the Adafactor-style gradient clipping~\citep{shazeer2018adafactor}. Table~\ref{tab:summary} of the Appendix lists the optimizer hyperparameters.
\\
The learning rate follows a Warmup Stable Decay (trapezoidal) schedule~\citep{hu2024minicpm}, known for its robustness to learning rate variations~\citep{hagele2024scaling}. We use a linear warmup over 10\% of the first epoch, followed by 80\% at peak LR, and a linear decay to 1\% of the maximum. Unlike one-cycle schedules that reset every epoch, our cyclic trapezoidal variant allows multiple cooldown phases across epochs, particularly beneficial for EEG training where masked token sampling introduces variability.
We apply Megatron-style initialization~\citep{megatron2019} with a standard deviation of $0.02$ for all transformer layers and the mask token, ensuring stable dynamics. Other parameters use PyTorch's default initializations.

A key factor for the success of foundation models is the simultaneous scaling of both training datasets and model architectures~\citep{touvron2023llama1}. We describe our scaling methodology to maximize computational efficiency and accommodate larger models within constrained resources.
To scale model capacity, we adjust depth, width, and number of attention heads while maintaining a fixed FFN ratio. Table~\ref{tab:config_summary} of the Appendix summarizes these configurations. This scaling strategy enables efficient capacity expansion while preserving architectural consistency across model sizes.
\\
Recent advances in NLP provide strong theoretical and empirical evidence for the existence of scaling laws~\citep{kaplan2020scaling, hoffmann2022training}, which govern the relationship between model size, training dynamics, optimization and initialization hyperparameters. We follow the power law $\eta \propto D^{\alpha_D}$, with $\alpha_D = -0.90$ and $D$ the model dimension, for the learning rate, as derived in~\citet{everett24scaling}. The optimal LR is first swept on the small model and then scaled accordingly.
\\
To efficiently train models, we use data parallelism, maintaining a constant batch size by reducing per-GPU loads for large models. A load-aware data-shuffling strategy groups samples by electrode count, shuffles within and across buckets, and balances batches across GPUs to avoid bottlenecks, for constant optimization steps and maximized throughput.
\\
Although scaling laws exist for adjusting AdamW momentum terms~\citep{malladi2022sdes}, our use of a constant effective batch size across models allows us to fix $\beta_1$ and $\beta_2$. Regarding initialization, while~\citet{hagele2024scaling} suggests scaling $\sigma_{\text{init}} \propto D^{-0.5}$, our width increase (from $200$ to $1,216$) leads us to keep $\sigma_{\text{init}} = 0.02$ fixed across scales.

\subsection{Downstream tasks}

\paragraph{Downstream task datasets}
To evaluate the performance and generalizability of our EEG foundation model, we perform extensive assessments across 10 diverse downstream tasks, selected to ensure comparability with existing models in the field. These tasks span a variety of EEG-based applications, including sleep staging, emotion and event classification, detection of stress and mental disorder , across the following datasets: 
PhysioNet-MI~\citep{goldberger2000physiobank}, 
BCIC-IV-2a~\citep{tangermann2012review}, 
TUEV~\citep{obeid2016temple},
TUAB~\citep{obeid2016temple}, 
HMC~\citep{alvarez2021inter}, 
ISRUC~\citep{khalighi2016isruc}, 
FACED~\citep{chen2023large}, 
Mumtaz~\citep{mumtaz2016mdd}, 
Mental Arithmetic (MAT)~\citep{zyma2019electroencephalograms}, 
and BCI2020-IV-3~\citep{jeong20222020}. 
A summary of these datasets is provided in Table~\ref{tab:downstream}, with a more detailed description available in the Appendix.
\begin{table}[h]
\caption{Overview of downstream tasks and datasets.}
\centering
    \small
    \begin{tabular}{ccccccc} 
        \toprule
        \textbf{Task}&\textbf{Dataset}&\textbf{\# Channels}&\textbf{Duration}&\textbf{\# Samples}&\textbf{Rate}&\textbf{\# Classes} \\
        \midrule
        Motor Imagery&PhysioNet-MI&64&4s&9,837&160Hz&4 \\
        &BCIC-IV-2a&22&4s&5,184&250Hz&4 \\
        Event Type&TUEV&16&5s&112,491&256Hz&6 \\
        Abnormal detection&TUAB&16&10s&409,455&256Hz&2 \\
 Sleep staging& HMC& 4& 30s& 137,243&256 Hz&5 \\
 & ISRUC& 6& 30s& 89,240&200Hz&5 \\
Emotion recognition& FACED& 32& 10s& 10,332&250Hz&9 \\
Mental disorder & Mumtaz& 19& 5s& 7,143&256Hz&2 \\
Mental stress& MAT& 20& 5s& 1,707&500Hz&2 \\
Imagined speech& BCIC2020-3& 64& 3s& 6,000 &256Hz&5 \\
    \bottomrule
    \end{tabular}
    \label{tab:downstream}
\end{table}

Our evaluation process maintains strict consistency with prior works by adhering to the same train/val/test splits used in earlier studies, ensuring that our results are directly comparable to baseline models. Specifically, we follow the protocols from CBraMod~\citep{wang2024cbramod}, LaBraM~\citep{jiang2024labram}, and BIOT~\citep{yang2024biot}, guaranteeing fair comparisons across tasks. For fairness in preprocessing, we adopt the same pipeline as the baselines. A notable correction was made for the ISRUC dataset, where we identified and removed a bug in the baseline code involving the inclusion of a chin electrode instead of an EEG electrode. Our results for REVE exclude the chin electrode, aligning with proper electrode placement.

\paragraph{Finetuning}
Fine-tuning EEG-based models presents unique challenges due to the small size of available datasets and the high noise levels in EEG recordings. Unlike large-scale vision datasets, EEG datasets are often limited in size, subject-dependent, and prone to distribution shifts across different recording setups. Effective fine-tuning must therefore maximize generalization while mitigating the risk of overfitting. To address this, we adopt a two-step fine-tuning strategy, incorporating techniques specifically designed to enhance stability and adaptability. This includes the use of parameter-efficient fine-tuning techniques~\citep{suzumura2024graph} tailored to this domain.

For downstream classification tasks, the two-step strategy, inspired by~\citet{kumar2022finetuning}, goes as follow: We first train a linear probe while keeping the encoder frozen, aligning the classifier with the pretrained feature space. Next, we unfreeze the encoder and fine-tune the entire network for task-specific adaptation, preserving the robustness of the pretrained model. Importantly, this two-step strategy is implemented as a single continuous training run,  where the backbone is initially frozen (i.e., only the head is trained) and later unfrozen. This approach is well-suited for EEG data, where distributions can shift significantly across datasets. We employ dropout and Mixup~\citep{zhang2018mixup} as data augmentation for improved robustness.
To further mitigate catastrophic forgetting and improve efficiency, we integrate Low-Rank Adaptation (LoRA) into the attention blocks, within the query, key, value, and output (QKVO) projection layers. Instead of fine-tuning the entire model, LoRA introduces trainable low-rank matrices that enable effective adaptation while preserving the integrity of the pretrained model’s knowledge~\citep{hu2022lora}. 

Each training step includes a warmup phase~\citep{kalra2024warmup} followed by a cooldown phase. The cooldown phase employs a Reduce-on-Plateau learning rate scheduler, which dynamically lowers the learning rate when training convergence slows to preventing overfitting.

To further enhance robustness, we explore model souping~\citep{wortsman2022soup}, which averages the weights of multiple fine-tuning runs to improve accuracy. Given the stochasticity and noise inherent in EEG datasets, souping smooths gradients and reduces variance across different fine-tuning trajectories. Our experiments confirm that this approach enhances generalization and produces more stable performance across diverse EEG tasks.

By integrating structured fine-tuning with data augmentation, LoRA and model souping, our approach effectively addresses the small-scale and noisy nature of EEG datasets. These techniques effectively ensure robust and generalized adaptation to downstream tasks.

\section{Results and Discussion}

We evaluate REVE against non-foundation and foundation model baselines on the previously discussed datasets.

\textbf{Non-Foundation Models:} We compare to EEGNet~\citep{lawhern2018eegnet}, EEGConformer~\citep{song2022eeg}, SPaRCNet~\citep{jing2023development}, ContraWR~\citep{yang2021self}, CNN-Transformer~\citep{peh2022transformer}, FFCL~\citep{li2022motor}, and ST-Transformer~\citep{song2021transformer}.

\textbf{Foundation Models:} We compare to BIOT~\citep{yang2024biot}, LaBraM~\citep{jiang2024labram} and CBraMod~\citep{wang2024cbramod}. We report results displayed in existing studies.


We report the balanced accuracy for each dataset and provide additional evaluation metrics in the appendix.

\begin{table}[!h]
    \centering
    \caption{Balanced accuracy ($\pm$ std) of different methods across 9 EEG classification task}
    \resizebox{\textwidth}{!}{%
    \begin{tabular}{lccccc}
        \toprule
         \textbf{Methods}&  TUAB&  TUEV&  PhysioNetMI&  BCI-IV-2a& FACED\\
         \midrule
         EEGNet&  0.7642 $\pm$ 0.0036&  0.3876 $\pm$ 0.0143&  0.5814 $\pm$ 0.0125&  0.4482 $\pm$ 0.0094& 0.4090 $\pm$ 0.0122\\
         EEGConformer&  0.7758 $\pm$ 0.0049&  0.4074 $\pm$ 0.0164&  0.6049 $\pm$ 0.0104&  0.4696 $\pm$ 0.0106& 0.4559 $\pm$ 0.0125\\
         SPaRCNet&  0.7896 $\pm$ 0.0018&  0.4161 $\pm$ 0.0262&  0.5932 $\pm$ 0.0152&  0.4635 $\pm$ 0.0117& 0.4673 $\pm$ 0.0155\\
         ContraWR&  0.7746 $\pm$ 0.0041&  0.4384 $\pm$ 0.0349&  0.5892 $\pm$ 0.0133&  0.4678 $\pm$ 0.0125& 0.4887 $\pm$ 0.0078\\
         CNN-Transformer&  0.7777 $\pm$ 0.0022&  0.4087 $\pm$ 0.0161&  0.6053 $\pm$ 0.0118&  0.4600 $\pm$ 0.0108& 0.4697 $\pm$ 0.0132\\
         FFCL&  0.7848 $\pm$ 0.0038&  0.3979 $\pm$ 0.0104&  0.5726 $\pm$ 0.0092&  0.4470 $\pm$ 0.0143& 0.4673 $\pm$ 0.0158\\
         ST-Transformer&  0.7966 $\pm$ 0.0023&  0.3984 $\pm$ 0.0228&  0.6035 $\pm$ 0.0081&  0.4575 $\pm$ 0.0145& 0.4810 $\pm$ 0.0079\\
         \midrule
         BIOT&  0.7959 $\pm$ 0.0057&  0.5281 $\pm$ 0.0225&  0.6153 $\pm$ 0.0154&  0.4748 $\pm$ 0.0093& 0.5118 $\pm$ 0.0118\\
         LaBraM-Base&  0.8140 $\pm$ 0.0019&  0.6409 $\pm$ 0.0065&  0.6173 $\pm$ 0.0122&  0.4869 $\pm$ 0.0085& 0.5273 $\pm$ 0.0107\\
         CbraMod&  0.8289 $\pm$ 0.0022&  0.6671 $\pm$ 0.0107&  0.6417 $\pm$ 0.0091&  0.5138 $\pm$ 0.0066& 0.5509 $\pm$ 0.0089\\
         \midrule
         REVE-Base&  \textbf{0.8315} $\pm$ 0.0014&  \textbf{0.6759} $\pm$ 0.0229&  \textbf{0.6480} $\pm$ 0.0140&  \textbf{0.6396} $\pm$ 0.0095& \textbf{0.5646} $\pm$ 0.0164\\
         \midrule
         &  ISRUC&  Mumtaz&  MAT&  BCI-2020-3& \textbf{Average}\\
         \midrule
         EEGNet&  0.7154 $\pm$ 0.0121&  0.9232 $\pm$ 0.0104&  0.6770 $\pm$ 0.0116&  0.4413 $\pm$ 0.0096& 0.5941 $\pm$ 0.0037 \\
         EEGConformer&  0.7400 $\pm$ 0.0133&  0.9308 $\pm$ 0.0117&  0.6805 $\pm$ 0.0123&  0.4506 $\pm$ 0.0133& 0.6128 $\pm$ 0.0044 \\
         SPaRCNet&  0.7487 $\pm$ 0.0075&  0.9316 $\pm$ 0.0095&  0.6879 $\pm$ 0.0107&  0.4426 $\pm$ 0.0156& 0.6156 $\pm$ 0.0047 \\
         ContraWR&  0.7402 $\pm$ 0.0126&  0.9195 $\pm$ 0.0115&  0.6631 $\pm$ 0.0097&  0.4257 $\pm$ 0.0162& 0.6119 $\pm$ 0.0053 \\
         CNN-Transformer&  0.7363 $\pm$ 0.0087&  0.9305 $\pm$ 0.0068&  0.6779 $\pm$ 0.0268&  0.4533 $\pm$ 0.0092& 0.6133 $\pm$ 0.0045 \\
         FFCL&  0.7277 $\pm$ 0.0182&  0.9314 $\pm$ 0.0038&  0.6798 $\pm$ 0.0142&  0.4678 $\pm$ 0.0197& 0.6085$\pm$ 0.0044 \\
         ST-Transformer&  0.7381 $\pm$ 0.0205&  0.9135 $\pm$ 0.0103&  0.6631 $\pm$ 0.0173&  0.4126 $\pm$ 0.0122& 0.6071 $\pm$ 0.0048 \\
         \midrule
         BIOT&  0.7527 $\pm$ 0.0121&  0.9358 $\pm$ 0.0052&  0.6875 $\pm$ 0.0186&  0.4920 $\pm$ 0.0086& 0.6438 $\pm$ 0.0044\\
         LaBraM-Base&  0.7633 $\pm$ 0.0102&  0.9409 $\pm$ 0.0079&  0.6909 $\pm$ 0.0125&  0.5060 $\pm$ 0.0155& 0.6653 $\pm$ 0.0031 \\
         CBraMod&  \textbf{0.7865} $\pm$ 0.0110&  0.9560 $\pm$ 0.0056&  0.7256 $\pm$ 0.0132&  0.5373 $\pm$ 0.0108& 0.6898 $\pm$ 0.0031\\
         \midrule
         REVE-Base&  0.7819 $\pm$ 0.0078\tablefootnote{NB: our preprocessing pipeline is different from the baseline and fixes a potential bug}&  \textbf{0.9644} $\pm$ 0.0097&  \textbf{0.7660} $\pm$ 0.0355&  \textbf{0.5635} $\pm$ 0.0123& \textbf{0.7150} $\pm$ 0.0057 \\
         \bottomrule
    \end{tabular}
    }
    \label{tab:merged}
\end{table}

\begin{table}[ht]
	\centering
    \caption{Impact of pretraining (PT) and weight freezing on REVE and baselines for PhysioNet-MI}
	\small
	\begin{tabular}{lccc} 
		\toprule
		&\multicolumn{3}{c}{\textbf{PhysioNet-MI, 4-class}}\\
    \cmidrule(lr){2-4}
        \textbf{Settings}&\textbf{Balanced Accuracy}&\textbf{Cohen’s Kappa}&\textbf{Weighted F1}\\
        \midrule
        CBraMod (w/ PT)& 0.6417 $\pm$ 0.0091& 0.5222 $\pm$ 0.0169&0.6427 $\pm$ 0.0100\\
		BIOT (w/ PT)& 0.6153 $\pm$ 0.0154&0.4875 $\pm$ 0.0272&0.6158 $\pm$ 0.0197\\
		LaBraM-Base (w/ PT)&0.6173 $\pm$ 0.0122&0.4912 $\pm$ 0.0192&0.6177 $\pm$ 0.0141\\
		REVE-Base (w/ PT)& \textbf{0.6480} $\pm$ 0.0140&\textbf{0.5306} $\pm$ 0.0187&\textbf{0.6484} $\pm$ 0.0170\\

        \midrule
        CBraMod (w/o PT)& \underline{0.6196} $\pm$ 0.0143& \underline{0.4994} $\pm$ 0.0289& \underline{0.6289}  $\pm$ 0.0179\\
        REVE-Base (w/o PT)& 0.5409 $\pm$ 0.0094& 0.3879 $\pm$ 0.0125&0.5421 $\pm$ 0.0101\\
        \midrule
		Cbramod (Frozen)&0.3845 $\pm$ 0.0345&0.2983 $\pm$ 0.0498 & 0.3946 $\pm$ 0.0378\\
		BIOT (Frozen)&0.3698 $\pm$ 0.0318&0.2703 $\pm$ 0.0472&0.3723 $\pm$ 0.0364\\
		LaBraM (Frozen)&0.3715 $\pm$ 0.0458&0.2814 $\pm$ 0.0586&0.3796 $\pm$ 0.0472\\
        REVE-Base (Frozen)& \underline{0.5371} $\pm$ 0.0052& \underline{0.3827} $\pm$ 0.0070& \underline{0.5376} $\pm$ 0.0033\\
		\bottomrule
	\end{tabular}

	\label{tab:result_FT}
\end{table}

\begin{table}[ht]
\centering
\caption{\textbf{Linear probing} results on downstream tasks for REVE and CBraMod models with (Pool) and without pooling across multiple EEG downstream tasks. Best results are highlighted in bold.  To ensure a fair comparison, we reproduced CBraMod~\citep{wang2024cbramod} using their official code and pretrained checkpoint, carefully following their classification pipeline (notably, no pooling) and matched architectural details to avoid any bias.}
  \resizebox{\textwidth}{!}{%
\begin{tabular}{lcccccc}
\toprule
\textbf{Dataset} & \textbf{REVE-B (Pool)} & \textbf{REVE-B} & \textbf{REVE-L (Pool)} & \textbf{REVE-L} & \textbf{CBraMod (Pool)} & \textbf{CBraMod} \\
\midrule
Mumtaz & 0.962 ± 0.003 & 0.931 ± 0.021 & \textbf{0.985 ± 0.006} & 0.980 ± 0.009 & 0.859 ± 0.009 & 0.907 ± 0.027 \\
M. Arithmetic & 0.725 ± 0.010 & \textbf{0.740 ± 0.073} & 0.712 ± 0.008 & 0.665 ± 0.103 & 0.500 ± 0.000 & 0.605 ± 0.020 \\
TUAB & 0.810 ± 0.007 & 0.809 ± 0.004 & \textbf{0.821 ± 0.004} & 0.809 ± 0.004 & 0.500 ± 0.000 & 0.500 ± 0.000 \\
PhysioNetMI & 0.537 ± 0.005 & 0.510 ± 0.012 & 0.551 ± 0.001 & \textbf{0.617 ± 0.000} & 0.256 ± 0.002 & 0.531 ± 0.015 \\
BCIC-IV-2a & 0.432 ± 0.004 & 0.517 ± 0.015 & 0.534 ± 0.001 & \textbf{0.603 ± 0.011} & 0.287 ± 0.023 & 0.376 ± 0.006 \\
ISRUC & 0.697 ± 0.011 & 0.662 ± 0.030 & 0.743 ± 0.004 & \textbf{0.758 ± 0.001} & 0.407 ± 0.049 & 0.430 ± 0.043 \\
HMC & 0.647 ± 0.008 & 0.604 ± 0.008 & 0.703 ± 0.003 & \textbf{0.710 ± 0.007} & 0.368 ± 0.001 & 0.538 ± 0.009 \\
BCIC2020-3 & 0.234 ± 0.009 & \textbf{0.390 ± 0.017} & 0.274 ± 0.001 & 0.378 ± 0.021 & 0.214 ± 0.003 & 0.374 ± 0.007 \\
TUEV & 0.592 ± 0.008 & 0.508 ± 0.073 & \textbf{0.630 ± 0.003} & 0.550 ± 0.014 & 0.219 ± 0.009 & 0.482 ± 0.037 \\
Faced & 0.240 ± 0.010 & 0.422 ± 0.028 & 0.283 ± 0.003 & \textbf{0.469 ± 0.007} & 0.117 ± 0.005 & 0.261 ± 0.013 \\
\midrule
\textbf{Avg.} & 0.586 & 0.609 & 0.623 & \textbf{0.654} & 0.373 & 0.501 \\
\bottomrule
\end{tabular}}
\label{tab:reve_LP_results}
\end{table}

Table~\ref{tab:merged} shows that REVE achieves state-of-the-art performance on the downstream tasks in this study, with an average gain of $2.5\%$, compared to CBraMod the highest performing baseline.
The results on ISRUC and HMC (Appendix~\ref{app:sleep}) show that the model effectively generalizes beyond the 10-second segments it was pretrained on, performing well on tasks with 30-second inputs, which highlights the strength of our positional encoding method. The results on TUEV highlight the model's ability to generalize to unseen electrode configurations, including bipolar setups never encountered during training. 

In addition to the detailed evaluation metrics provided in Appendix~\ref{app:detailedresults}, we report the performance of the Large model across our downstream tasks in Table~\ref{tab:reve_LP_results}. We observe that the Large model consistently produces richer embeddings, leading to improved linear probing performance compared to the Base model. Model souping consistently improved performance, averaging a 1.5\% gain when combining at least 5 Base or Large models. For example, REVE-Base achieved 69.6\% balanced accuracy on TUEV using the 10 models from Table~\ref{tab:merged}. However, souping showed limited benefits for the small models and sometimes led to negative outcomes.

Table~\ref{tab:result_FT} highlights the importance of REVE’s pretraining phase. Without pretraining, CBraMod outperforms REVE by at least 8\%. However, pretraining improves REVE-Base by 11\%, while CBraMod gains only 2\%, a trend also observed in the LaBraM paper. This suggests that REVE benefits more significantly from pretraining, whereas other models derive most of their performance from architectural design rather than pretraining learned representations. A key advantage of REVE is its ability to produce high-quality latent spaces without heavy fine-tuning, as evidenced by linear probing results in Table~\ref{tab:reve_LP_results}: REVE consistently outperforms CBraMod across all downstream tasks and model sizes, with REVE-Large achieving nearly 17\% higher performance. These results also highlight REVE’s ability to scale effectively with model size, yielding richer and more generalizable embeddings as capacity increases. Providing rich, ready-to-use embeddings is crucial for enabling zero-shot analysis, faster BCI calibration, and improved performance in low-data or sparsely annotated settings. REVE also benefits from its spatial encoding strategy, which enables transfer across diverse EEG configurations. In Appendix~\ref{app:ablationssl}, we further demonstrate the contribution of our secondary loss function, a novel component of our framework, which proves particularly effective in frozen-feature scenarios. The secondary objective reconstructs masked tokens using a compressed, global representation from attention pooling. This pooling acts as an information bottleneck, forcing the model to distill key information from the entire input sequence into a single vector. As shown by Table~\ref{tab:second_loss}, the secondary loss mainly improves the quality of the frozen embeddings of the model.

\section{Limitations and Future Work}

The model has some limitations, requiring signals to be at least one second and multiples of one second. A way to address this could be to leverage padding with causal masking. 

While the focus has been on collecting large EEG datasets for pretraining, an important next step could be to curate this data more selectively. This includes removing low-quality recordings, balancing distributions, and identifying representative subsets, especially given the inherently noisy nature of EEG signals. Our current pretraining corpus aggregates 92 publicly available EEG datasets spanning over 25,000 subjects, which helps reduce overfitting to any single source. However, most public EEG data originates from North America and Europe, resulting in limited demographic diversity—a key limitation that calls for broader, more equitable data collection efforts. To partially mitigate such imbalances, we leverage self-supervised learning (MAE), which has been shown to be robust to long-tailed and heterogeneous data distributions~\citep{xu2023learning}. Targeted selection strategies, combined with robust SSL objectives, could help focus on the most informative and complementary data for building stronger, fairer, and more efficient foundation models. Thanks to its flexibility in handling any EEG configuration, REVE could itself guide this curation process.

We also plan to extend our study to diverse tasks, including zero-/few-shot regimes. This first iteration uses a simple MAE approach and a standard transformer, but future improvements could leverage more advanced SSL techniques and  architectures. We release the model’s code, weights and guidelines for adapting it to mainstream EEG tasks. In parallel, our findings point toward the presence of scaling effects in EEG foundation models. Identifying precise scaling laws that capture how model size, data volume, and downstream performance interact would be valuable for future work.

\section{Conclusion}

EEG research has lacked a foundation model that transfers robustly across devices, montages, and tasks—especially under linear probing. REVE contributes to bridging this gap. Trained on ~60,000 hours from 92 datasets and ~25,000 subjects, REVE combines a 4D Fourier positional encoding that natively supports arbitrary electrode layouts and sequence lengths with masked autoencoding enhanced by spatio-temporal block masking and a global-token secondary loss.
Across 10 benchmarks, it sets a new state of the art (average +2.5\% balanced accuracy over prior foundation models), delivers up to ~17\% gains in linear probing, and generalizes to unseen/bipolar montages and longer inputs than used in pretraining. These properties enable faster BCI calibration, more reliable cross-site clinical deployment, and standardized embeddings for downstream analytics. We release code, weights, loaders for arbitrary 3D coordinates, and training/eval recipes. We invite the community to extend REVE to broader populations and modalities (MEG/iEEG/OPM-MEG), and to co-build a cross-montage benchmark for fair, scalable EEG evaluation.

\section{Acknowledgments}

This research was supported by the French National Research Agency (ANR) through its AI@IMT program and grant ANR-24-CE23-7365, as well as by a grant from the Brittany region. Further support was provided by a Discovery Grant from the Natural Sciences and Engineering Research Council of Canada (NSERC), by funding from the Canada Research Chairs program and the Fonds de recherche du Québec – Nature et technologies (FRQ-NT). This work was granted access to the HPC resources of IDRIS under the allocation 2024-AD011015237R1 made by GENCI, as well as HPC provided by Digital Alliance Canada. 

\vfill
\pagebreak

\bibliography{references}

\begin{thebibliography}{151}
\providecommand{\natexlab}[1]{#1}
\providecommand{\url}[1]{\texttt{#1}}
\expandafter\ifx\csname urlstyle\endcsname\relax
  \providecommand{\doi}[1]{doi: #1}\else
  \providecommand{\doi}{doi: \begingroup \urlstyle{rm}\Url}\fi

\bibitem[Accou et~al.(2023)Accou, Bollens, Gillis, Verheijen, Van~hamme, and Francart]{accou2023sparrkulee}
Bernd Accou, Lies Bollens, Marlies Gillis, Wendy Verheijen, Hugo Van~hamme, and Tom Francart.
\newblock {SparrKULee: A Speech-Evoked Auditory Response Repository of the KU Leuven, Containing EEG of 85 Participants}.
\newblock \emph{BioRxiv}, pages 2023--07, 2023.

\bibitem[Achiam et~al.(2023)Achiam, Adler, Agarwal, Ahmad, Akkaya, Aleman, Almeida, Altenschmidt, Altman, Anadkat, et~al.]{achiam2023gpt4}
Josh Achiam, Steven Adler, Sandhini Agarwal, Lama Ahmad, Ilge Akkaya, Florencia~Leoni Aleman, Diogo Almeida, Janko Altenschmidt, Sam Altman, Shyamal Anadkat, et~al.
\newblock {GPT-4 Technical Report}.
\newblock \emph{ArXiv Preprint ArXiv:2303.08774}, 2023.

\bibitem[Aguado-Lopez et~al.(2024)Aguado-Lopez, Palenciano, Penalver, Diaz-Gutierrez, Lopez-Garcia, Avancini, Ciria, and Ruz]{ds005089:1.0.1}
Blanca Aguado-Lopez, Ana~F. Palenciano, Jose M.~G. Penalver, Paloma Diaz-Gutierrez, David Lopez-Garcia, Chiara Avancini, Luis~F. Ciria, and Maria Ruz.
\newblock {"Proactive Selective Attention Across Competition Contexts"}, 2024.

\bibitem[Alexander et~al.(2017)Alexander, Escalera, Ai, Andreotti, Febre, Mangone, Vega-Potler, Langer, Alexander, Kovacs, et~al.]{alexander2017open}
Lindsay~M Alexander, Jasmine Escalera, Lei Ai, Charissa Andreotti, Karina Febre, Alexander Mangone, Natan Vega-Potler, Nicolas Langer, Alexis Alexander, Meagan Kovacs, et~al.
\newblock {An Open Resource for Transdiagnostic Research in Pediatric Mental Health and Learning Disorders}.
\newblock \emph{Scientific Data}, 4\penalty0 (1):\penalty0 1--26, 2017.

\bibitem[Alkin et~al.(2024)Alkin, Miklautz, Hochreiter, and Brandstetter]{alkin2024mim}
Benedikt Alkin, Lukas Miklautz, Sepp Hochreiter, and Johannes Brandstetter.
\newblock {MIM-Refiner: A Contrastive Learning Boost from Intermediate Pre-Trained Representations}.
\newblock \emph{ArXiv Preprint ArXiv:2402.10093}, 2024.

\bibitem[Alvarez-Estevez and Rijsman(2021)]{alvarez2021inter}
Diego Alvarez-Estevez and Roselyne~M Rijsman.
\newblock {Inter-Database Validation of a Deep Learning Approach for Automatic Sleep Scoring}.
\newblock \emph{PLOS One}, 16\penalty0 (8):\penalty0 e0256111, 2021.

\bibitem[Amorim et~al.(2023)Amorim, Zheng, Lee, Herman, Ghassemi, Sivaraju, Gaspard, Hofmeijer, van Putten, Reyna, et~al.]{amorim2023care}
Edilberto Amorim, Wei-Long Zheng, Jong~Woo Lee, Susan Herman, Mohammad Ghassemi, Adithya Sivaraju, Nicolas Gaspard, Jeannette Hofmeijer, Michel~JAM van Putten, Matthew Reyna, et~al.
\newblock {I-CARE: International Cardiac Arrest Research Consortium Database}.
\newblock \emph{https://physionet.org/content/i-care/2.0}, 2023.

\bibitem[Araya et~al.(2023)Araya, Mendez-Orellana, and Rodriguez-Fernandez]{ds004279:1.1.1}
Carlos~Valle Araya, Carolina Mendez-Orellana, and Maria Rodriguez-Fernandez.
\newblock {"Large Spanish EEG"}, 2023.

\bibitem[Aric{\`o} et~al.(2014)Aric{\`o}, Aloise, Schettini, Salinari, Mattia, and Cincotti]{BNCI2014009}
Pietro Aric{\`o}, F~Aloise, Francesca Schettini, Serenella Salinari, D~Mattia, and Febo Cincotti.
\newblock {Influence of P300 Latency Jitter on Event Related Potential-Based Brain--Computer Interface Performance}.
\newblock \emph{Journal of Neural Engineering}, 11\penalty0 (3):\penalty0 035008, 2014.

\bibitem[Aristimunha et~al.(2023)Aristimunha, Carrara, Guetschel, Sedlar, Rodrigues, Sosulski, Narayanan, Bjareholt, Quentin, Schirrmeister, Kalunga, Darmet, Gregoire, Abdul~Hussain, Gatti, Goncharenko, Thielen, Moreau, Roy, Jayaram, Barachant, and Chevallier]{Aristimunha_Mother_of_all_2023}
Bruno Aristimunha, Igor Carrara, Pierre Guetschel, Sara Sedlar, Pedro Rodrigues, Jan Sosulski, Divyesh Narayanan, Erik Bjareholt, Barthelemy Quentin, Robin~Tibor Schirrmeister, Emmanuel Kalunga, Ludovic Darmet, Cattan Gregoire, Ali Abdul~Hussain, Ramiro Gatti, Vladislav Goncharenko, Jordy Thielen, Thomas Moreau, Yannick Roy, Vinay Jayaram, Alexandre Barachant, and Sylvain Chevallier.
\newblock {{Mother of All BCI Benchmarks}}, 2023.
\newblock URL \url{https://moabb.neurotechx.com/docs/index.html}.

\bibitem[Bai et~al.(2023)Bai, Bai, Chu, Cui, Dang, Deng, Fan, Ge, Han, Huang, et~al.]{bai2023qwen}
Jinze Bai, Shuai Bai, Yunfei Chu, Zeyu Cui, Kai Dang, Xiaodong Deng, Yang Fan, Wenbin Ge, Yu~Han, Fei Huang, et~al.
\newblock {Qwen Technical Report}.
\newblock \emph{ArXiv Preprint ArXiv:2309.16609}, 2023.

\bibitem[Bajwa1 et~al.(2024)Bajwa1, Nilsen1, René~Skukies1, Aamodt1, Ernst2, Storm1, and Bjørn E.~Juel1]{ds005620:1.0.0}
Imad~J. Bajwa1, Andre~S. Nilsen1, 3~René~Skukies1, Arnfinn Aamodt1, Gernot Ernst2, Johan~F. Storm1, and 2~Bjørn E.~Juel1.
\newblock {"A Repeated Awakening Study Exploring the Capacity of Complexity Measures to Capture Dreaming During Propofol Sedation"}, 2024.

\bibitem[Barachant(2012)]{AlexMI}
Alexandre Barachant.
\newblock \emph{{Commande Robuste d'un Effecteur par une Interface Cerveau Machine EEG Asynchrone}}.
\newblock PhD thesis, Universit{\'e} de Grenoble, 2012.

\bibitem[Baykan and Schütz(2024)]{ds005586:2.0.0}
Cemre Baykan and Alexander~C. Schütz.
\newblock {"Electroencephalographic Responses to the Number of Objects in Partially Occluded and Uncovered Scenes"}, 2024.

\bibitem[Bialas et~al.(2023)Bialas, Teoh, Anderson, and Lalor]{ds004408:1.0.0}
Ole Bialas, Emily Teoh, Andrew Anderson, and Edmund Lalor.
\newblock {"Invariant Encoding of Phonemes in Neural Responses to Continuous Speech"}, 2023.

\bibitem[Caron et~al.(2021)Caron, Touvron, Misra, J{\'e}gou, Mairal, Bojanowski, and Joulin]{caron2021emerging}
Mathilde Caron, Hugo Touvron, Ishan Misra, Herv{\'e} J{\'e}gou, Julien Mairal, Piotr Bojanowski, and Armand Joulin.
\newblock {Emerging Properties in Self-Supervised Vision Transformers}.
\newblock In \emph{Proceedings of the IEEE/CVF International Conference on Computer Vision}, pages 9650--9660, 2021.

\bibitem[Cavanagh and Jackson(2022)]{ds004315:1.0.0}
James~F Cavanagh and Trevor C~J Jackson.
\newblock {"Mood Manipulation and PST, Experiment 1"}, 2022.

\bibitem[Chacón and Wriessnegger(2023)]{ds004324:1.0.0}
Luis Alberto~Barradas Chacón and Selina~C. Wriessnegger.
\newblock {"Toonfaces"}, 2023.

\bibitem[Chen et~al.(2023)Chen, Wang, Huang, Hu, Shen, and Zhang]{chen2023large}
Jingjing Chen, Xiaobin Wang, Chen Huang, Xin Hu, Xinke Shen, and Dan Zhang.
\newblock {A Large Finer-Grained Affective Computing EEG Dataset}.
\newblock \emph{Scientific Data}, 10\penalty0 (1):\penalty0 740, 2023.

\bibitem[Chien et~al.(2022)Chien, Goh, Sandino, and Cheng]{chien2022maeeg}
Hsiang-Yun~Sherry Chien, Hanlin Goh, Christopher~Michael Sandino, and Joseph~Yitan Cheng.
\newblock {MaEEG: Masked Auto-Encoder for EEG Representation Learning}.
\newblock In \emph{NeurIPS 2022 Workshop on Learning from Time Series for Health}, 2022.

\bibitem[Cho et~al.(2017)Cho, Ahn, Ahn, Kwon, and Jun]{Cho2017}
Hohyun Cho, Minkyu Ahn, Sangtae Ahn, Moonyoung Kwon, and Sung~Chan Jun.
\newblock {EEG Datasets for Motor Imagery Brain--Computer Interface}.
\newblock \emph{Gigascience}, 6\penalty0 (7):\penalty0 gix034, 2017.

\bibitem[Chowdhery et~al.(2023)Chowdhery, Narang, Devlin, Bosma, Mishra, Roberts, Barham, Chung, Sutton, Gehrmann, et~al.]{chowdhery2023palm}
Aakanksha Chowdhery, Sharan Narang, Jacob Devlin, Maarten Bosma, Gaurav Mishra, Adam Roberts, Paul Barham, Hyung~Won Chung, Charles Sutton, Sebastian Gehrmann, et~al.
\newblock {PaLM: Scaling Language Modeling with Pathways}.
\newblock \emph{Journal of Machine Learning Research}, 24\penalty0 (240):\penalty0 1--113, 2023.

\bibitem[Cordoba-Silva et~al.(2023)Cordoba-Silva, Maya, Valderrama, Giraldo, Betancourt-Zapata, Salgado-Vascob, Marín-Sánchez, Gómez-Ortega, and Ettenberger]{ds004840:1.0.1}
Jose Cordoba-Silva, Rafael Maya, Mario Valderrama, Luis~Felipe Giraldo, William Betancourt-Zapata, Andrés Salgado-Vascob, Juliana Marín-Sánchez, Viviana Gómez-Ortega, and Mark Ettenberger.
\newblock {"Dataset of Electrophysiological Signals (EEG, ECG, EMG) During Music Therapy with Adult Burn Patients in the Intensive Care Unit"}, 2023.

\bibitem[Cui et~al.(2024)Cui, Jeong, Th{\"o}lke, Medani, Jerbi, Joshi, and Leahy]{cui2024neuro}
Wenhui Cui, Woojae Jeong, Philipp Th{\"o}lke, Takfarinas Medani, Karim Jerbi, Anand~A Joshi, and Richard~M Leahy.
\newblock {Neuro-GPT: Towards a Foundation Model for EEG}.
\newblock In \emph{2024 IEEE International Symposium on Biomedical Imaging (ISBI)}, pages 1--5. IEEE, 2024.

\bibitem[Daly et~al.(2020)Daly, Nicolaou, Williams, Hwang, Kirke, Miranda, and Nasuto]{ds002721:1.0.1}
Ian Daly, Nicoletta Nicolaou, Duncan Williams, Faustina Hwang, Alexis Kirke, Eduardo Miranda, and Slawomir~J. Nasuto.
\newblock {"An EEG Dataset Recorded During Affective Music Listening"}, 2020.

\bibitem[Dao(2024)]{dao2023flashattention2}
Tri Dao.
\newblock {FlashAttention-2: Faster Attention with Better Parallelism and Work Partitioning}.
\newblock In \emph{International Conference on Learning Representations (ICLR)}, 2024.

\bibitem[Dayma et~al.(2021)Dayma, Patil, Cuenca, Saifullah, Abraham, Lê~Khac, Melas, and Ghosh]{Dayma_DALLE_Mini_2021}
Boris Dayma, Suraj Patil, Pedro Cuenca, Khalid Saifullah, Tanishq Abraham, Phúc Lê~Khac, Luke Melas, and Ritobrata Ghosh.
\newblock {Dall·E Mini}, 7 2021.
\newblock URL \url{https://github.com/borisdayma/dalle-mini}.

\bibitem[D{\'e}fossez et~al.(2023)D{\'e}fossez, Caucheteux, Rapin, Kabeli, and King]{defossez2023decoding}
Alexandre D{\'e}fossez, Charlotte Caucheteux, J{\'e}r{\'e}my Rapin, Ori Kabeli, and Jean-R{\'e}mi King.
\newblock {Decoding Speech Perception from Non-Invasive Brain Recordings}.
\newblock \emph{Nature Machine Intelligence}, 5\penalty0 (10):\penalty0 1097--1107, 2023.

\bibitem[Delorme and Braboszcz(2021)]{ds003969:1.0.0}
Arnaud Delorme and Claire Braboszcz.
\newblock {"Meditation Vs Thinking Task"}, 2021.

\bibitem[Delorme and Brandmeyer(2024)]{ds001787:1.1.1}
Arnaud Delorme and Tracy Brandmeyer.
\newblock {"EEG Meditation Study"}, 2024.

\bibitem[Delorme and Fabre-Thorpe(2020)]{ds002680:1.0.0}
Arnaud Delorme and Michele Fabre-Thorpe.
\newblock {"Go-Nogo Categorization and Detection Task"}, 2020.

\bibitem[Detti(2020)]{detti2020siena}
Paolo Detti.
\newblock {Siena Scalp EEG Database}.
\newblock \emph{Physionet. Doi}, 10:\penalty0 493, 2020.

\bibitem[Detti et~al.(2020)Detti, Vatti, and Zabalo Manrique~de Lara]{detti2020eeg}
Paolo Detti, Giampaolo Vatti, and Garazi Zabalo Manrique~de Lara.
\newblock {EEG Synchronization Analysis for Seizure Prediction: A Study on Data of Noninvasive Recordings}.
\newblock \emph{Processes}, 8\penalty0 (7):\penalty0 846, 2020.

\bibitem[Dreyer et~al.(2023)Dreyer, Roc, Pillette, Rimbert, and Lotte]{dreyer2023large}
Pauline Dreyer, Aline Roc, L{\'e}a Pillette, S{\'e}bastien Rimbert, and Fabien Lotte.
\newblock {A Large EEG Database with Users’ Profile Information for Motor Imagery Brain-Computer Interface Research}.
\newblock \emph{Scientific Data}, 10\penalty0 (1):\penalty0 580, 2023.

\bibitem[Dubey et~al.(2024)Dubey, Jauhri, Pandey, Kadian, Al-Dahle, Letman, Mathur, Schelten, Yang, Fan, et~al.]{dubey2024llama3}
Abhimanyu Dubey, Abhinav Jauhri, Abhinav Pandey, Abhishek Kadian, Ahmad Al-Dahle, Aiesha Letman, Akhil Mathur, Alan Schelten, Amy Yang, Angela Fan, et~al.
\newblock {The Llama 3 Herd of Models}.
\newblock \emph{ArXiv Preprint ArXiv:2407.21783}, 2024.

\bibitem[El~Ouahidi et~al.(2024)El~Ouahidi, Gripon, Pasdeloup, Bouallegue, Farrugia, and Lioi]{el2024strong}
Yassine El~Ouahidi, Vincent Gripon, Bastien Pasdeloup, Ghaith Bouallegue, Nicolas Farrugia, and Giulia Lioi.
\newblock {A Strong and Simple Deep Learning Baseline for BCI Motor Imagery Decoding}.
\newblock \emph{IEEE Transactions on Neural Systems and Rehabilitation Engineering}, 2024.

\bibitem[Esteban et~al.(2024)Esteban, Chica, and González-López]{ds005273:1.0.0}
Pablo Rodríguez-San Esteban, Ana~B. Chica, and José~A. González-López.
\newblock {"Neural Representation of Consciously Seen and Unseen Information"}, 2024.

\bibitem[Everett et~al.(2024)Everett, Xiao, Wortsman, Alemi, Novak, Liu, Gur, Sohl-Dickstein, Kaelbling, Lee, and Pennington]{everett24scaling}
Katie~E Everett, Lechao Xiao, Mitchell Wortsman, Alexander~A Alemi, Roman Novak, Peter~J Liu, Izzeddin Gur, Jascha Sohl-Dickstein, Leslie~Pack Kaelbling, Jaehoon Lee, and Jeffrey Pennington.
\newblock {Scaling Exponents Across Parameterizations and Optimizers}.
\newblock In Ruslan Salakhutdinov, Zico Kolter, Katherine Heller, Adrian Weller, Nuria Oliver, Jonathan Scarlett, and Felix Berkenkamp, editors, \emph{Proceedings of the 41St International Conference on Machine Learning}, volume 235 of \emph{Proceedings of Machine Learning Research}, pages 12666--12700. PMLR, 21--27 Jul 2024.
\newblock URL \url{https://proceedings.mlr.press/v235/everett24a.html}.

\bibitem[Faller et~al.(2012)Faller, Vidaurre, Solis-Escalante, Neuper, and Scherer]{BNCI2015001}
Josef Faller, Carmen Vidaurre, Teodoro Solis-Escalante, Christa Neuper, and Reinhold Scherer.
\newblock {Autocalibration and Recurrent Adaptation: Towards a Plug and Play Online ERD-BCI}.
\newblock \emph{IEEE Transactions on Neural Systems and Rehabilitation Engineering}, 20\penalty0 (3):\penalty0 313--319, 2012.

\bibitem[Gehrke et~al.(2024)Gehrke, Akman, Chen, Lopes, and Gramann]{ds003846:2.0.2}
Lukas Gehrke, Sezen Akman, Albert Chen, Pedro Lopes, and Klaus Gramann.
\newblock {"Prediction Error"}, 2024.

\bibitem[Geiping and Goldstein(2023)]{geiping2023cramming}
Jonas Geiping and Tom Goldstein.
\newblock {Cramming: Training a Language Model on a Single GPU in One Day.}
\newblock In \emph{International Conference on Machine Learning}, pages 11117--11143. PMLR, 2023.

\bibitem[Gifford et~al.(2022)Gifford, Dwivedi, Roig, and Cichy]{gifford2022large}
Alessandro~T Gifford, Kshitij Dwivedi, Gemma Roig, and Radoslaw~M Cichy.
\newblock {A Large and Rich EEG Dataset for Modeling Human Visual Object Recognition}.
\newblock \emph{NeuroImage}, 264:\penalty0 119754, 2022.

\bibitem[Goldberger et~al.(2000)Goldberger, Amaral, Glass, Hausdorff, Ivanov, Mark, Mietus, Moody, Peng, and Stanley]{goldberger2000physiobank}
Ary~L Goldberger, Luis~AN Amaral, Leon Glass, Jeffrey~M Hausdorff, Plamen~Ch Ivanov, Roger~G Mark, Joseph~E Mietus, George~B Moody, Chung-Kang Peng, and H~Eugene Stanley.
\newblock {Physiobank, Physiotoolkit, and Physionet: Components of a New Research Resource for Complex Physiologic Signals}.
\newblock \emph{Circulation}, 101\penalty0 (23):\penalty0 e215--e220, 2000.

\bibitem[Grootswagers et~al.(2022)Grootswagers, Zhou, Robinson, Hebart, and Carlson]{ds003825:1.2.0}
Tijl Grootswagers, Ivy Zhou, Amanda Robinson, Martin Hebart, and Thomas Carlson.
\newblock {"Human Electroencephalography Recordings from 50 Subjects for 22,248 Images from 1,854 Object Concepts"}, 2022.

\bibitem[Grootswagers et~al.(2023{\natexlab{a}})Grootswagers, Robinson, Shatek, and Carlson]{ds004816:1.0.1}
Tijl Grootswagers, Amanda Robinson, Sofia Shatek, and Thomas Carlson.
\newblock {"EEG-attention-rsvp-exp1"}, 2023{\natexlab{a}}.

\bibitem[Grootswagers et~al.(2023{\natexlab{b}})Grootswagers, Robinson, Shatek, and Carlson]{ds004817:1.0.1}
Tijl Grootswagers, Amanda Robinson, Sofia Shatek, and Thomas Carlson.
\newblock {"EEG-attention-rsvp-exp2"}, 2023{\natexlab{b}}.

\bibitem[Grootswagers et~al.(2024)Grootswagers, Robinson, Shatek, and Carlson]{ds004357:1.0.1}
Tijl Grootswagers, Amanda Robinson, Sofia Shatek, and Thomas Carlson.
\newblock {"Features-EEG"}, 2024.

\bibitem[Guetschel et~al.(2024)Guetschel, Moreau, and Tangermann]{guetschel2024s}
Pierre Guetschel, Thomas Moreau, and Michael Tangermann.
\newblock {S-JEPA: Towards Seamless Cross-Dataset Transfer Through Dynamic Spatial Attention}.
\newblock \emph{ArXiv Preprint ArXiv:2403.11772}, 2024.

\bibitem[Guger et~al.(2009)Guger, Daban, Sellers, Holzner, Krausz, Carabalona, Gramatica, and Edlinger]{BNCI2015003}
Christoph Guger, Shahab Daban, Eric Sellers, Clemens Holzner, Gunther Krausz, Roberta Carabalona, Furio Gramatica, and Guenter Edlinger.
\newblock {How Many People Are Able To Control a P300-Based Brain--Computer Interface (BCI)?}
\newblock \emph{Neuroscience Letters}, 462\penalty0 (1):\penalty0 94--98, 2009.

\bibitem[H{\"a}gele et~al.(2024)H{\"a}gele, Bakouch, Kosson, Allal, Von~Werra, and Jaggi]{hagele2024scaling}
Alexander H{\"a}gele, Elie Bakouch, Atli Kosson, Loubna~Ben Allal, Leandro Von~Werra, and Martin Jaggi.
\newblock {Scaling Laws and Compute-Optimal Training Beyond Fixed Training Durations}.
\newblock \emph{ArXiv Preprint ArXiv:2405.18392}, 2024.

\bibitem[Hassall et~al.(2022{\natexlab{a}})Hassall, Yan, and Hunt]{ds004152:1.1.2}
Cameron~D. Hassall, Yan Yan, and Laurence~T. Hunt.
\newblock {"Drum Trainer"}, 2022{\natexlab{a}}.

\bibitem[Hassall et~al.(2022{\natexlab{b}})Hassall, Yan, and Hunt]{ds004264:1.1.0}
Cameron~D. Hassall, Yan Yan, and Laurence~T. Hunt.
\newblock {"Steer the Ship"}, 2022{\natexlab{b}}.

\bibitem[Hassall et~al.(2024)Hassall, Hunt, and Holroyd]{ds004147:1.0.2}
Cameron~D. Hassall, Laurence~T. Hunt, and Clay~B. Holroyd.
\newblock {"Average Task Value"}, 2024.

\bibitem[Hatlestad-Hall et~al.(2022)Hatlestad-Hall, Rygvold, and Andersson]{ds003775:1.2.1}
Christoffer Hatlestad-Hall, Trine~Waage Rygvold, and Stein Andersson.
\newblock {"SRM Resting-State EEG"}, 2022.

\bibitem[Haupt et~al.(2024)Haupt, Graumann, Teng, Kaltenbach, and Cichy]{ds004951:1.0.0}
Marleen Haupt, Monika Graumann, Santani Teng, Carina Kaltenbach, and Radoslaw~M. Cichy.
\newblock {"Braille Letters - EEG"}, 2024.

\bibitem[He and Wu(2019)]{he2019transfer}
He~He and Dongrui Wu.
\newblock {Transfer Learning for Brain--Computer Interfaces: A Euclidean Space Data Alignment Approach}.
\newblock \emph{IEEE Transactions on Biomedical Engineering}, 67\penalty0 (2):\penalty0 399--410, 2019.

\bibitem[He et~al.(2022)He, Chen, Xie, Li, Doll{\'a}r, and Girshick]{he2022masked}
Kaiming He, Xinlei Chen, Saining Xie, Yanghao Li, Piotr Doll{\'a}r, and Ross Girshick.
\newblock {Masked Autoencoders Are Scalable Vision Learners}.
\newblock In \emph{Proceedings of the IEEE/CVF Conference on Computer Vision and Pattern Recognition}, pages 16000--16009, 2022.

\bibitem[Helbing et~al.(2024)Helbing, Draschkow, and Võ]{ds005189:1.0.1}
Jason Helbing, Dejan Draschkow, and Melissa L.-H. Võ.
\newblock {"Search Superiority Recollection Familiarity"}, 2024.

\bibitem[Hendrycks and Gimpel(2016)]{hendrycks2016gaussian}
Dan Hendrycks and Kevin Gimpel.
\newblock {Gaussian Error Linear Units (GELUs)}.
\newblock \emph{ArXiv Preprint ArXiv:1606.08415}, 2016.

\bibitem[Hoffmann et~al.(2022)Hoffmann, Borgeaud, Mensch, Buchatskaya, Cai, Rutherford, Casas, Hendricks, Welbl, Clark, et~al.]{hoffmann2022training}
Jordan Hoffmann, Sebastian Borgeaud, Arthur Mensch, Elena Buchatskaya, Trevor Cai, Eliza Rutherford, Diego de~Las Casas, Lisa~Anne Hendricks, Johannes Welbl, Aidan Clark, et~al.
\newblock {Training Compute-Optimal Large Language Models}.
\newblock \emph{ArXiv Preprint ArXiv:2203.15556}, 2022.

\bibitem[Hoffmann et~al.(2008)Hoffmann, Vesin, Ebrahimi, and Diserens]{EPFLP300}
Ulrich Hoffmann, Jean-Marc Vesin, Touradj Ebrahimi, and Karin Diserens.
\newblock {An Efficient P300-Based Brain--Computer Interface for Disabled Subjects}.
\newblock \emph{Journal of Neuroscience Methods}, 167\penalty0 (1):\penalty0 115--125, 2008.

\bibitem[Hu et~al.(2022)Hu, Shen, Wallis, Allen-Zhu, Li, Wang, Wang, and Chen]{hu2022lora}
Edward~J Hu, Yelong Shen, Phillip Wallis, Zeyuan Allen-Zhu, Yuanzhi Li, Shean Wang, Lu~Wang, and Weizhu Chen.
\newblock {LoRA: Low-Rank Adaptation of Large Language Models}.
\newblock In \emph{International Conference on Learning Representations}, 2022.
\newblock URL \url{https://openreview.net/forum?id=nZeVKeeFYf9}.

\bibitem[Hu et~al.(2024)Hu, Tu, Han, Cui, He, Zhao, Long, Zheng, Fang, Huang, Zhang, Thai, Wang, Yao, Zhao, Zhou, Cai, Zhai, Ding, Jia, Zeng, dahai li, Liu, and Sun]{hu2024minicpm}
Shengding Hu, Yuge Tu, Xu~Han, Ganqu Cui, Chaoqun He, Weilin Zhao, Xiang Long, Zhi Zheng, Yewei Fang, Yuxiang Huang, Xinrong Zhang, Zhen~Leng Thai, Chongyi Wang, Yuan Yao, Chenyang Zhao, Jie Zhou, Jie Cai, Zhongwu Zhai, Ning Ding, Chao Jia, Guoyang Zeng, dahai li, Zhiyuan Liu, and Maosong Sun.
\newblock {MiniCPM: Unveiling the Potential of Small Language Models with Scalable Training Strategies}.
\newblock In \emph{First Conference on Language Modeling}, 2024.
\newblock URL \url{https://openreview.net/forum?id=3X2L2TFr0f}.

\bibitem[Jeong et~al.(2022)Jeong, Cho, Lee, Lee, Shin, Kweon, Mill{\'a}n, M{\"u}ller, and Lee]{jeong20222020}
Ji-Hoon Jeong, Jeong-Hyun Cho, Young-Eun Lee, Seo-Hyun Lee, Gi-Hwan Shin, Young-Seok Kweon, Jos{\'e} del~R Mill{\'a}n, Klaus-Robert M{\"u}ller, and Seong-Whan Lee.
\newblock {2020 International Brain--Computer Interface Competition: A Review}.
\newblock \emph{Frontiers in Human Neuroscience}, 16:\penalty0 898300, 2022.

\bibitem[Jiang et~al.(2023)Jiang, Sablayrolles, Mensch, Bamford, Chaplot, Casas, Bressand, Lengyel, Lample, Saulnier, et~al.]{jiang2023mistral}
Albert~Q Jiang, Alexandre Sablayrolles, Arthur Mensch, Chris Bamford, Devendra~Singh Chaplot, Diego de~las Casas, Florian Bressand, Gianna Lengyel, Guillaume Lample, Lucile Saulnier, et~al.
\newblock {Mistral 7B}.
\newblock \emph{ArXiv Preprint ArXiv:2310.06825}, 2023.

\bibitem[Jiang et~al.(2024)Jiang, Zhao, and liang Lu]{jiang2024labram}
Weibang Jiang, Liming Zhao, and Bao liang Lu.
\newblock {Large Brain Model for Learning Generic Representations with Tremendous EEG Data in BCI}.
\newblock In \emph{The Twelfth International Conference on Learning Representations}, 2024.

\bibitem[Jing et~al.(2023)Jing, Ge, Hong, Fernandes, Lin, Yang, An, Struck, Herlopian, Karakis, et~al.]{jing2023development}
Jin Jing, Wendong Ge, Shenda Hong, Marta~Bento Fernandes, Zhen Lin, Chaoqi Yang, Sungtae An, Aaron~F Struck, Aline Herlopian, Ioannis Karakis, et~al.
\newblock {Development of Expert-Level Classification of Seizures and Rhythmic and Periodic Patterns During EEG Interpretation}.
\newblock \emph{Neurology}, 100\penalty0 (17):\penalty0 e1750--e1762, 2023.

\bibitem[Kahana et~al.(2023)Kahana, Rudoler, Lohnas, Healey, Aka, Broitman, Crutchley, Crutchley, Alm, Katerman, Miller, Kuhn, Li, Long, Miller, Paron, Pazdera, Pedisich, and Weidemann]{ds004395:2.0.0}
Michael~J. Kahana, Joseph~H. Rudoler, Lynn~J. Lohnas, Karl Healey, Ada Aka, Adam Broitman, Elizabeth Crutchley, Patrick Crutchley, Kylie~H. Alm, Brandon~S. Katerman, Nicole~E. Miller, Joel~R. Kuhn, Yuxuan Li, Nicole~M. Long, Jonathan Miller, Madison~D. Paron, Jesse~K. Pazdera, Isaac Pedisich, and Christoph~T. Weidemann.
\newblock {"Penn Electrophysiology of Encoding and Retrieval Study (Peers)"}, 2023.

\bibitem[Kalra and Barkeshli(2024)]{kalra2024warmup}
Dayal~Singh Kalra and Maissam Barkeshli.
\newblock {Why Warmup the Learning Rate? Underlying Mechanisms and Improvements}.
\newblock \emph{ArXiv Preprint ArXiv:2406.09405}, 2024.

\bibitem[Kalunga et~al.(2015)Kalunga, Chevallier, and Barth{\'e}lemy]{Kalunga2016}
Emmanuel~K Kalunga, Sylvain Chevallier, and Quentin Barth{\'e}lemy.
\newblock {Using Riemannian Geometry for SSVEP-Based Brain Computer Interface}.
\newblock \emph{ArXiv Preprint ArXiv:1501.03227}, 2015.

\bibitem[Kaplan et~al.(2020)Kaplan, McCandlish, Henighan, Brown, Chess, Child, Gray, Radford, Wu, and Amodei]{kaplan2020scaling}
Jared Kaplan, Sam McCandlish, Tom Henighan, Tom~B Brown, Benjamin Chess, Rewon Child, Scott Gray, Alec Radford, Jeffrey Wu, and Dario Amodei.
\newblock {Scaling Laws for Neural Language Models}.
\newblock \emph{ArXiv Preprint ArXiv:2001.08361}, 2020.

\bibitem[Kekecs and Farahzadi(2024)]{ds004572:1.2.0}
Zoltan Kekecs and Yeganeh Farahzadi.
\newblock {"OTKA PLB-HYP Study1"}, 2024.

\bibitem[Khalighi et~al.(2016)Khalighi, Sousa, Santos, and Nunes]{khalighi2016isruc}
Sirvan Khalighi, Teresa Sousa, Jos{\'e}~Moutinho Santos, and Urbano Nunes.
\newblock {Isruc-Sleep: A Comprehensive Public Dataset for Sleep Researchers}.
\newblock \emph{Computer Methods and Programs in Biomedicine}, 124:\penalty0 180--192, 2016.

\bibitem[Khan et~al.(2022)Khan, Ul~Ain, Kamboh, Butt, Shafait, Alamgir, Stricker, and Shafait]{khan2022nmt}
Hassan~Aqeel Khan, Rahat Ul~Ain, Awais~Mehmood Kamboh, Hammad~Tanveer Butt, Saima Shafait, Wasim Alamgir, Didier Stricker, and Faisal Shafait.
\newblock {The NMT Scalp EEG Dataset: An Open-Source Annotated Dataset of Healthy and Pathological EEG Recordings for Predictive Modeling}.
\newblock \emph{Frontiers in Neuroscience}, 15:\penalty0 755817, 2022.

\bibitem[Kirillov et~al.(2023)Kirillov, Mintun, Ravi, Mao, Rolland, Gustafson, Xiao, Whitehead, Berg, Lo, et~al.]{kirillov2023segment}
Alexander Kirillov, Eric Mintun, Nikhila Ravi, Hanzi Mao, Chloe Rolland, Laura Gustafson, Tete Xiao, Spencer Whitehead, Alexander~C Berg, Wan-Yen Lo, et~al.
\newblock {Segment Anything}.
\newblock In \emph{Proceedings of the IEEE/CVF International Conference on Computer Vision}, pages 4015--4026, 2023.

\bibitem[Korczowski et~al.(2019{\natexlab{a}})Korczowski, Cederhout, Andreev, Cattan, Rodrigues, Gautheret, and Congedo]{BI2015a}
Louis Korczowski, Martine Cederhout, Anton Andreev, Gr{\'e}goire Cattan, Pedro Luiz~Coelho Rodrigues, Violette Gautheret, and Marco Congedo.
\newblock \emph{{Brain Invaders Calibration-Less P300-Based BCI With Modulation of Flash Duration Dataset (Bi2015A)}}.
\newblock PhD thesis, GIPSA-lab, 2019{\natexlab{a}}.

\bibitem[Korczowski et~al.(2019{\natexlab{b}})Korczowski, Ostaschenko, Andreev, Cattan, Rodrigues, Gautheret, and Congedo]{BI2014a}
Louis Korczowski, Ekaterina Ostaschenko, Anton Andreev, Gr{\'e}goire Cattan, Pedro Luiz~Coelho Rodrigues, Violette Gautheret, and Marco Congedo.
\newblock \emph{{Brain Invaders Calibration-Less P300-Based BCI Using Dry EEG Electrodes Dataset (Bi2014A)}}.
\newblock PhD thesis, GIPSA-lab, 2019{\natexlab{b}}.

\bibitem[Korczowski et~al.(2019{\natexlab{c}})Korczowski, Ostaschenko, Andreev, Cattan, Rodrigues, Gautheret, and Congedo]{BI2014b}
Louis Korczowski, Ekaterina Ostaschenko, Anton Andreev, Gr{\'e}goire Cattan, Pedro Luiz~Coelho Rodrigues, Violette Gautheret, and Marco Congedo.
\newblock \emph{{Brain Invaders Solo Versus Collaboration: Multi-User P300-Based Brain-Computer Interface Dataset (Bi2014B)}}.
\newblock PhD thesis, GIPSA-lab, 2019{\natexlab{c}}.

\bibitem[Kumar et~al.(2022)Kumar, Raghunathan, Jones, Ma, and Liang]{kumar2022finetuning}
Ananya Kumar, Aditi Raghunathan, Robbie~Matthew Jones, Tengyu Ma, and Percy Liang.
\newblock {Fine-Tuning Can Distort Pretrained Features and Underperform Out-Of-Distribution}.
\newblock In \emph{International Conference on Learning Representations}, 2022.
\newblock URL \url{https://openreview.net/forum?id=UYneFzXSJWh}.

\bibitem[Lawhern et~al.(2018)Lawhern, Solon, Waytowich, Gordon, Hung, and Lance]{lawhern2018eegnet}
Vernon~J Lawhern, Amelia~J Solon, Nicholas~R Waytowich, Stephen~M Gordon, Chou~P Hung, and Brent~J Lance.
\newblock {EEGnet: A Compact Convolutional Neural Network for EEG-Based Brain--Computer Interfaces}.
\newblock \emph{Journal of Neural Engineering}, 15\penalty0 (5):\penalty0 056013, 2018.

\bibitem[Lee et~al.(2019)Lee, Kwon, Kim, Kim, Lee, Williamson, Fazli, and Lee]{Lee2019}
Min-Ho Lee, O-Yeon Kwon, Yong-Jeong Kim, Hong-Kyung Kim, Young-Eun Lee, John Williamson, Siamac Fazli, and Seong-Whan Lee.
\newblock {EEG Dataset and OpenBMI Toolbox for Three BCI Paradigms: An Investigation Into BCI Illiteracy}.
\newblock \emph{Gigascience}, 8\penalty0 (5):\penalty0 giz002, 2019.

\bibitem[Leeb et~al.(2007)Leeb, Lee, Keinrath, Scherer, Bischof, and Pfurtscheller]{BNCI2014004}
Robert Leeb, Felix Lee, Claudia Keinrath, Reinhold Scherer, Horst Bischof, and Gert Pfurtscheller.
\newblock {Brain--Computer Communication: Motivation, Aim, and Impact of Exploring a Virtual Apartment}.
\newblock \emph{IEEE Transactions on Neural Systems and Rehabilitation Engineering}, 15\penalty0 (4):\penalty0 473--482, 2007.

\bibitem[Lei~Ba et~al.(2016)Lei~Ba, Kiros, and Hinton]{lei2016layernorm}
Jimmy Lei~Ba, Jamie~Ryan Kiros, and Geoffrey~E Hinton.
\newblock {Layer Normalization}.
\newblock \emph{ArXiv e-Prints}, pages ArXiv--1607, 2016.

\bibitem[Li et~al.(2022)Li, Ding, Zhang, and Xiu]{li2022motor}
Hongli Li, Man Ding, Ronghua Zhang, and Chunbo Xiu.
\newblock {Motor Imagery EEG Classification Algorithm Based on CNN-LSTM Feature Fusion Network}.
\newblock \emph{Biomedical Signal Processing and Control}, 72:\penalty0 103342, 2022.

\bibitem[Li and Zhao(2024)]{ds005697:1.0.0}
Weilong Li and Jiaxin Zhao.
\newblock {"PerceiveImagine"}, 2024.

\bibitem[Liu et~al.(2024)Liu, Wei, Wang, Lv, Duan, Li, Zhao, Wang, Chen, Shi, et~al.]{Liu2024}
Haijie Liu, Penghu Wei, Haochong Wang, Xiaodong Lv, Wei Duan, Meijie Li, Yan Zhao, Qingmei Wang, Xinyuan Chen, Gaige Shi, et~al.
\newblock {An EEG Motor Imagery Dataset for Brain Computer Interface in Acute Stroke Patients}.
\newblock \emph{Scientific Data}, 11\penalty0 (1):\penalty0 131, 2024.

\bibitem[Lotte et~al.(2018)Lotte, Bougrain, Cichocki, Clerc, Congedo, Rakotomamonjy, and Yger]{lotte2018review}
Fabien Lotte, Laurent Bougrain, Andrzej Cichocki, Maureen Clerc, Marco Congedo, Alain Rakotomamonjy, and Florian Yger.
\newblock {A Review of Classification Algorithms for EEG-Based Brain--Computer Interfaces: A 10 Year Update}.
\newblock \emph{Journal of Neural Engineering}, 15\penalty0 (3):\penalty0 031005, 2018.

\bibitem[Lowe et~al.(2023)Lowe, Robinson, Yamamoto, Hogendoorn, and Johnston]{ds004603:1.1.0}
Benjamin Lowe, Jonathan Robinson, Naohide Yamamoto, Hinze Hogendoorn, and Patrick Johnston.
\newblock {"Visual Attribute-Specific Contextual Trajectory Paradigm"}, 2023.

\bibitem[Makowski et~al.(2023)Makowski, Te, Kirk, and Ngoi]{ds004582:1.0.0}
Dominique Makowski, An-Shu Te, Stephanie Kirk, and Zi~Liang Ngoi.
\newblock {"FakeFaceEmo Data"}, 2023.

\bibitem[Malladi et~al.(2022)Malladi, Lyu, Panigrahi, and Arora]{malladi2022sdes}
Sadhika Malladi, Kaifeng Lyu, Abhishek Panigrahi, and Sanjeev Arora.
\newblock {On the SDEs and Scaling Rules for Adaptive Gradient Algorithms}.
\newblock \emph{Advances in Neural Information Processing Systems}, 35:\penalty0 7697--7711, 2022.

\bibitem[Markiewicz et~al.(2021)Markiewicz, Gorgolewski, Feingold, Blair, Halchenko, Miller, Hardcastle, Wexler, Esteban, Goncavles, et~al.]{markiewicz2021openneuro}
Christopher~J Markiewicz, Krzysztof~J Gorgolewski, Franklin Feingold, Ross Blair, Yaroslav~O Halchenko, Eric Miller, Nell Hardcastle, Joe Wexler, Oscar Esteban, Mathias Goncavles, et~al.
\newblock {The OpenNeuro Resource for Sharing of Neuroscience Data}.
\newblock \emph{eLife}, 10:\penalty0 e71774, 2021.

\bibitem[Metwalli et~al.(2024)Metwalli, Ahmed, Emil, Radwan, Barakat, and Ahmed]{ds005262:1.0.0}
Donia Metwalli, Eslam Ahmed, Antony Emil, Yousef~A. Radwan, Mariam Barakat, and Anas Ahmed.
\newblock {"ArEEG: Arabic Inner Speech EEG Dataset"}, 2024.

\bibitem[Moerel et~al.(2022)Moerel, Grootswagers, Robinson, Shatek, Woolgar, Carlson, and Rich]{ds004043:1.0.0}
Denise Moerel, Tijl Grootswagers, Amanda Robinson, Sophia Shatek, Alexandra Woolgar, Thomas Carlson, and Anina Rich.
\newblock {"The Time-Course of Feature-Based Attention Effects Dissociated from Temporal Expectation and Target-Related Processes"}, 2022.

\bibitem[Mohammadi~Foumani et~al.(2024)Mohammadi~Foumani, Mackellar, Ghane, Irtza, Nguyen, and Salehi]{mohammadi2024eeg2rep}
Navid Mohammadi~Foumani, Geoffrey Mackellar, Soheila Ghane, Saad Irtza, Nam Nguyen, and Mahsa Salehi.
\newblock {EEG2Rep: Enhancing Self-Supervised EEG Representation Through Informative Masked Inputs}.
\newblock In \emph{Proceedings of the 30th ACM SIGKDD Conference on Knowledge Discovery and Data Mining}, pages 5544--5555, 2024.

\bibitem[Mumtaz(2016)]{mumtaz2016mdd}
Wajid Mumtaz.
\newblock {MDD Patients and Healthy Controls EEG Data (New)}.
\newblock \emph{Figshare, Dataset}, 2016.

\bibitem[Nakanishi et~al.(2015)Nakanishi, Wang, Wang, and Jung]{Nakanishi2015}
Masaki Nakanishi, Yijun Wang, Yu-Te Wang, and Tzyy-Ping Jung.
\newblock {A Comparison Study of Canonical Correlation Analysis Based Methods for Detecting Steady-State Visual Evoked Potentials}.
\newblock \emph{PLOS One}, 10\penalty0 (10):\penalty0 e0140703, 2015.

\bibitem[Obeid and Picone(2016)]{obeid2016temple}
Iyad Obeid and Joseph Picone.
\newblock {The Temple University Hospital EEG Data Corpus}.
\newblock \emph{Frontiers in Neuroscience}, 10:\penalty0 196, 2016.

\bibitem[Ofner et~al.(2017)Ofner, Schwarz, Pereira, and M{\"u}ller-Putz]{Ofner2017}
Patrick Ofner, Andreas Schwarz, Joana Pereira, and Gernot~R M{\"u}ller-Putz.
\newblock {Upper Limb Movements Can Be Decoded From the Time-Domain of Low-Frequency EEG}.
\newblock \emph{PLOS One}, 12\penalty0 (8):\penalty0 e0182578, 2017.

\bibitem[Onton and Makeig(2022)]{ds003004:1.1.1}
Julie Onton and Scott Makeig.
\newblock {"Imagined Emotion Study"}, 2022.

\bibitem[Oostenveld and Praamstra(2001)]{oostenveld2001five}
Robert Oostenveld and Peter Praamstra.
\newblock {The Five Percent Electrode System for High-Resolution EEG and ERP Measurements}.
\newblock \emph{Clinical Neurophysiology}, 112\penalty0 (4):\penalty0 713--719, 2001.

\bibitem[Papastylianou et~al.(2023)Papastylianou, Ramele, Citi, Cinel, and Poli]{ds004477:1.0.0}
Tasos Papastylianou, Rodrigo Ramele, Luca Citi, Caterina Cinel, and Riccardo Poli.
\newblock {"PES - Pandemic Emergency Scenario"}, 2023.

\bibitem[Peh et~al.(2022)Peh, Yao, and Dauwels]{peh2022transformer}
Wei~Yan Peh, Yuanyuan Yao, and Justin Dauwels.
\newblock {Transformer Convolutional Neural Networks for Automated Artifact Detection in Scalp EEG}.
\newblock In \emph{2022 44th Annual International Conference of the IEEE Engineering in Medicine \& Biology Society (EMBC)}, pages 3599--3602. IEEE, 2022.

\bibitem[Radford et~al.(2021)Radford, Kim, Hallacy, Ramesh, Goh, Agarwal, Sastry, Askell, Mishkin, Clark, et~al.]{radford2021learning}
Alec Radford, Jong~Wook Kim, Chris Hallacy, Aditya Ramesh, Gabriel Goh, Sandhini Agarwal, Girish Sastry, Amanda Askell, Pamela Mishkin, Jack Clark, et~al.
\newblock {Learning Transferable Visual Models from Natural Language Supervision}.
\newblock In \emph{International Conference on Machine Learning}, pages 8748--8763. PMLR, 2021.

\bibitem[Ram et~al.(2024)Ram, Ganesan, and Kiran]{ram2024harmful}
Shanker Ram, Sambhu Ganesan, and Yajat~Nagaraj Kiran.
\newblock {Harmful Brain Activity Classification of Spectrograms with Transfer Deep Learning}.
\newblock In \emph{2024 IEEE 7th International Conference on Multimedia Information Processing and Retrieval (MIPR)}, pages 499--502. IEEE, 2024.

\bibitem[Ribeiro and Castelo-Branco(2021)]{ds003690:1.0.0}
Maria~J. Ribeiro and Miguel Castelo-Branco.
\newblock {"EEG, ECG and Pupil Data from Young and Older Adults: Rest and Auditory Cued Reaction Time Tasks"}, 2021.

\bibitem[Riccio et~al.(2013)Riccio, Simione, Schettini, Pizzimenti, Inghilleri, Belardinelli, Mattia, and Cincotti]{BNCI2014008}
Angela Riccio, Luca Simione, Francesca Schettini, Alessia Pizzimenti, Maurizio Inghilleri, Marta~Olivetti Belardinelli, Donatella Mattia, and Febo Cincotti.
\newblock {Attention and P300-Based BCI Performance in People with Amyotrophic Lateral Sclerosis}.
\newblock \emph{Frontiers in Human Neuroscience}, 7:\penalty0 732, 2013.

\bibitem[Rockhill et~al.(2020)Rockhill, Jackson, George, Aron, and Swann]{ds002778:1.0.1}
Alexander~P. Rockhill, Nicko Jackson, Jobi George, Adam Aron, and Nicole~C. Swann.
\newblock {"UC San Diego Resting State EEG Data from Patients with Parkinson's Disease"}, 2020.

\bibitem[Rudoler et~al.(2023)Rudoler, Dougherty, Katerman, Bruska, Chang, Halpern, Diamond, and Kahana]{ds004706:1.0.0}
Joseph~H. Rudoler, Matthew~R. Dougherty, Brandon~S. Katerman, James~P. Bruska, Woohyeuk Chang, David~J. Halpern, Nicholas~B. Diamond, and Michael~J. Kahana.
\newblock {"Spatial Memory and Non-Invasive Closed-Loop Stimulus Timing"}, 2023.

\bibitem[Scherer et~al.(2015)Scherer, Faller, Friedrich, Opisso, Costa, K{\"u}bler, and M{\"u}ller-Putz]{BNCI2015004}
Reinhold Scherer, Josef Faller, Elisabeth~VC Friedrich, Eloy Opisso, Ursula Costa, Andrea K{\"u}bler, and Gernot~R M{\"u}ller-Putz.
\newblock {Individually Adapted Imagery Improves Brain-Computer Interface Performance in End-Users with Disability}.
\newblock \emph{PLOS One}, 10\penalty0 (5):\penalty0 e0123727, 2015.

\bibitem[Schirrmeister et~al.(2017)Schirrmeister, Springenberg, Fiederer, Glasstetter, Eggensperger, Tangermann, Hutter, Burgard, and Ball]{Schirrmeister2017}
Robin~Tibor Schirrmeister, Jost~Tobias Springenberg, Lukas Dominique~Josef Fiederer, Martin Glasstetter, Katharina Eggensperger, Michael Tangermann, Frank Hutter, Wolfram Burgard, and Tonio Ball.
\newblock {Deep Learning with Convolutional Neural Networks for EEG Decoding and Visualization}.
\newblock \emph{Human Brain Mapping}, 38\penalty0 (11):\penalty0 5391--5420, 2017.

\bibitem[Shan et~al.(2022)Shan, Cappelloni, and Maddox]{ds004356:1.0.0}
Tong Shan, Madeline~S. Cappelloni, and Ross~K. Maddox.
\newblock {"Music and Speech Elicit Similar Subcortical Responses in Human Listeners"}, 2022.

\bibitem[Shatek et~al.(2021)Shatek, Robinson, Grootswagers, and Carlson]{ds003885:1.0.0}
Sophia~M. Shatek, Amanda~K. Robinson, Tijl Grootswagers, and Thomas~A. Carlson.
\newblock {"Capacity for Movement Is a Major Organisational Principle in Object Representations: EEG Data from Experiment 2"}, 2021.

\bibitem[Shatek et~al.(2023)Shatek, Robinson, Grootswagers, and Carlson]{ds003887:1.2.3}
Sophia~M. Shatek, Amanda~K. Robinson, Tijl Grootswagers, and Thomas~A. Carlson.
\newblock {"Capacity for Movement Is an Organisational Principle in Object Representations: EEG Data from Experiment 2"}, 2023.

\bibitem[Shazeer(2020)]{shazeer2020glu}
Noam Shazeer.
\newblock {GLU Variants Improve Transformer}.
\newblock \emph{ArXiv Preprint ArXiv:2002.05202}, 2020.

\bibitem[Shazeer and Stern(2018)]{shazeer2018adafactor}
Noam Shazeer and Mitchell Stern.
\newblock {Adafactor: Adaptive Learning Rates with Sublinear Memory Cost}.
\newblock In \emph{International Conference on Machine Learning}, pages 4596--4604. PMLR, 2018.

\bibitem[Shin et~al.(2016)Shin, von L{\"u}hmann, Blankertz, Kim, Jeong, Hwang, and M{\"u}ller]{Shin2017A}
Jaeyoung Shin, Alexander von L{\"u}hmann, Benjamin Blankertz, Do-Won Kim, Jichai Jeong, Han-Jeong Hwang, and Klaus-Robert M{\"u}ller.
\newblock {Open Access Dataset for EEG+ Nirs Single-Trial Classification}.
\newblock \emph{IEEE Transactions on Neural Systems and Rehabilitation Engineering}, 25\penalty0 (10):\penalty0 1735--1745, 2016.

\bibitem[Shirazi et~al.(2024{\natexlab{a}})Shirazi, Franco, Hoffmann, Esper, Truong, Delorme, Milham, and Makeig]{ds005508:1.0.0}
Seyed~Yahya Shirazi, Alexandre Franco, Maurício~Scopel Hoffmann, Nathalia~B. Esper, Dung Truong, Arnaud Delorme, Michael Milham, and Scott Makeig.
\newblock {"Healthy Brain Network (HBN) EEG - Release 4"}, 2024{\natexlab{a}}.

\bibitem[Shirazi et~al.(2024{\natexlab{b}})Shirazi, Franco, Scopel~Hoffmann, Esper, Truong, Delorme, Milham, and Makeig]{shirazi2024hbn}
Seyed~Yahya Shirazi, Alexandre Franco, Maur{\'\i}cio Scopel~Hoffmann, Nathalia~B Esper, Dung Truong, Arnaud Delorme, Michael~P Milham, and Scott Makeig.
\newblock {HBN-EEG: The Fair Implementation of the Healthy Brain Network (HBN) Electroencephalography Dataset}.
\newblock \emph{BioRxiv}, pages 2024--10, 2024{\natexlab{b}}.

\bibitem[Shirazi et~al.(2025)Shirazi, Franco, Hoffmann, Esper, Truong, Delorme, Milham, and Makeig]{ds005509:1.0.1}
Seyed~Yahya Shirazi, Alexandre Franco, Maurício~Scopel Hoffmann, Nathalia~B. Esper, Dung Truong, Arnaud Delorme, Michael Milham, and Scott Makeig.
\newblock {"Healthy Brain Network (HBN) EEG - Release 5"}, 2025.

\bibitem[Shoeybi et~al.(2019)Shoeybi, Patwary, Puri, LeGresley, Casper, and Catanzaro]{megatron2019}
Mohammad Shoeybi, Mostofa Patwary, Raul Puri, Patrick LeGresley, Jared Casper, and Bryan Catanzaro.
\newblock {Megatron-LM: Training Multi-Billion Parameter Language Models Using Model Parallelism}.
\newblock \emph{ArXiv Preprint ArXiv:1909.08053}, 2019.

\bibitem[Siefert et~al.(2024)Siefert, Uppuluri, Mu, Tandoc, Antony, and Schapiro]{ds005121:1.0.1}
Elizabeth~M. Siefert, Sindhuja Uppuluri, Jianing Mu, Marlie~C. Tandoc, James~W. Antony, and Anna~C. Schapiro.
\newblock {"Siefert2024"}, 2024.

\bibitem[Song et~al.(2021)Song, Jia, Yang, and Xie]{song2021transformer}
Yonghao Song, Xueyu Jia, Lie Yang, and Longhan Xie.
\newblock {Transformer-Based Spatial-Temporal Feature Learning for EEG Decoding}.
\newblock \emph{ArXiv Preprint ArXiv:2106.11170}, 2021.

\bibitem[Song et~al.(2022)Song, Zheng, Liu, and Gao]{song2022eeg}
Yonghao Song, Qingqing Zheng, Bingchuan Liu, and Xiaorong Gao.
\newblock {EEG Conformer: Convolutional Transformer for EEG Decoding and Visualization}.
\newblock \emph{IEEE Transactions on Neural Systems and Rehabilitation Engineering}, 31:\penalty0 710--719, 2022.

\bibitem[Sosulski et~al.(2021)Sosulski, H{\"u}bner, Klein, and Tangermann]{Sosulski2019}
Jan Sosulski, David H{\"u}bner, Aaron Klein, and Michael Tangermann.
\newblock {Online Optimization of Stimulation Speed in an Auditory Brain-Computer Interface Under Time Constraints}.
\newblock \emph{ArXiv Preprint ArXiv:2109.06011}, 2021.

\bibitem[Suzumura et~al.(2024)Suzumura, Kanezashi, and Akahori]{suzumura2024graph}
Toyotaro Suzumura, Hiroki Kanezashi, and Shotaro Akahori.
\newblock {Graph Adapter of EEG Foundation Models for Parameter Efficient Fine Tuning}.
\newblock \emph{ArXiv Preprint ArXiv:2411.16155}, 2024.

\bibitem[Tangermann et~al.(2012)Tangermann, M{\"u}ller, Aertsen, Birbaumer, Braun, Brunner, Leeb, Mehring, Miller, M{\"u}ller-Putz, et~al.]{tangermann2012review}
Michael Tangermann, Klaus-Robert M{\"u}ller, Ad~Aertsen, Niels Birbaumer, Christoph Braun, Clemens Brunner, Robert Leeb, Carsten Mehring, Kai~J Miller, Gernot~R M{\"u}ller-Putz, et~al.
\newblock {Review of the BCI Competition IV}.
\newblock \emph{Frontiers in Neuroscience}, 6:\penalty0 55, 2012.

\bibitem[Taylor et~al.(2024)Taylor, Sinn, Iaia, and Fiebach]{ds005594:1.0.3}
Jack~E. Taylor, Rasmus Sinn, Cosimo Iaia, and Christian~J. Fiebach.
\newblock {"Alphabetic Decision Task (Arial Light Font)"}, 2024.

\bibitem[Touvron et~al.(2023)Touvron, Lavril, Izacard, Martinet, Lachaux, Lacroix, Rozi{\`e}re, Goyal, Hambro, Azhar, et~al.]{touvron2023llama1}
Hugo Touvron, Thibaut Lavril, Gautier Izacard, Xavier Martinet, Marie-Anne Lachaux, Timoth{\'e}e Lacroix, Baptiste Rozi{\`e}re, Naman Goyal, Eric Hambro, Faisal Azhar, et~al.
\newblock {Llama: Open and Efficient Foundation Language Models}.
\newblock \emph{ArXiv Preprint ArXiv:2302.13971}, 2023.

\bibitem[Van~Dijk et~al.(2022)Van~Dijk, Van~Wingen, Denys, Olbrich, Van~Ruth, and Arns]{van2022two}
Hanneke Van~Dijk, Guido Van~Wingen, Damiaan Denys, Sebastian Olbrich, Rosalinde Van~Ruth, and Martijn Arns.
\newblock {The Two Decades Brainclinics Research Archive for Insights in Neurophysiology (TDBRAIN) Database}.
\newblock \emph{Scientific Data}, 9\penalty0 (1):\penalty0 333, 2022.

\bibitem[Vaswani(2017)]{vaswani2017attention}
A~Vaswani.
\newblock {Attention Is All You Need}.
\newblock \emph{Advances in Neural Information Processing Systems}, 2017.

\bibitem[Veillette et~al.(2022)Veillette, Heald, Wittenbrink, and Nusbaum]{ds004284:1.0.0}
J.~Veillette, S.~Heald, B.~Wittenbrink, and H.~Nusbaum.
\newblock {"EEG-Neuroforecasting"}, 2022.

\bibitem[Veillette et~al.(2023)Veillette, Lopes, and Nusbaum]{ds004561:1.0.0}
John Veillette, Pedro Lopes, and Howard Nusbaum.
\newblock {"Illusion of Agency Over Electrically-Actuated Movements"}, 2023.

\bibitem[Wang et~al.(2024{\natexlab{a}})Wang, Liu, He, Xu, Ma, and Li]{wang2024eegpt}
Guangyu Wang, Wenchao Liu, Yuhong He, Cong Xu, Lin Ma, and Haifeng Li.
\newblock {EEGPT: Pretrained Transformer for Universal and Reliable Representation of EEG Signals}.
\newblock In \emph{The Thirty-Eighth Annual Conference on Neural Information Processing Systems}, 2024{\natexlab{a}}.

\bibitem[Wang et~al.(2024{\natexlab{b}})Wang, Zhao, Luo, Zhou, Jiang, Li, Li, and Pan]{wang2024cbramod}
Jiquan Wang, Sha Zhao, Zhiling Luo, Yangxuan Zhou, Haiteng Jiang, Shijian Li, Tao Li, and Gang Pan.
\newblock {CBraMod: A Criss-Cross Brain Foundation Model for EEG Decoding}.
\newblock \emph{ArXiv Preprint ArXiv:2412.07236}, 2024{\natexlab{b}}.

\bibitem[Wang et~al.(2022)Wang, Duan, Dong, Ding, and Lei]{ds004148:1.0.1}
Yulin Wang, Wei Duan, Debo Dong, Lihong Ding, and Xu~Lei.
\newblock {"A Test-Retest Resting and Cognitive State EEG Dataset"}, 2022.

\bibitem[Warner et~al.(2024)Warner, Chaffin, Clavi{\'e}, Weller, Hallstr{\"o}m, Taghadouini, Gallagher, Biswas, Ladhak, Aarsen, et~al.]{warner2024modernbert}
Benjamin Warner, Antoine Chaffin, Benjamin Clavi{\'e}, Orion Weller, Oskar Hallstr{\"o}m, Said Taghadouini, Alexis Gallagher, Raja Biswas, Faisal Ladhak, Tom Aarsen, et~al.
\newblock {Smarter, Better, Faster, Longer: A Modern Bidirectional Encoder for Fast, Memory Efficient, and Long Context Finetuning and Inference}.
\newblock \emph{ArXiv Preprint ArXiv:2412.13663}, 2024.

\bibitem[Wortsman et~al.(2022)Wortsman, Ilharco, Gadre, Roelofs, Gontijo-Lopes, Morcos, Namkoong, Farhadi, Carmon, Kornblith, and Schmidt]{wortsman2022soup}
Mitchell Wortsman, Gabriel Ilharco, Samir~Ya Gadre, Rebecca Roelofs, Raphael Gontijo-Lopes, Ari~S Morcos, Hongseok Namkoong, Ali Farhadi, Yair Carmon, Simon Kornblith, and Ludwig Schmidt.
\newblock {Model Soups: Averaging Weights of Multiple Fine-Tuned Models Improves Accuracy Without Increasing Inference Time}.
\newblock In Kamalika Chaudhuri, Stefanie Jegelka, Le~Song, Csaba Szepesvari, Gang Niu, and Sivan Sabato, editors, \emph{Proceedings of the 39th International Conference on Machine Learning}, volume 162 of \emph{Proceedings of Machine Learning Research}, pages 23965--23998. PMLR, 17--23 Jul 2022.
\newblock URL \url{https://proceedings.mlr.press/v162/wortsman22a.html}.

\bibitem[Wortsman et~al.(2023)Wortsman, Dettmers, Zettlemoyer, Morcos, Farhadi, and Schmidt]{wortsman2023stable}
Mitchell Wortsman, Tim Dettmers, Luke Zettlemoyer, Ari Morcos, Ali Farhadi, and Ludwig Schmidt.
\newblock {Stable and Low-Precision Training for Large-Scale Vision-Language Models}.
\newblock \emph{Advances in Neural Information Processing Systems}, 36:\penalty0 10271--10298, 2023.

\bibitem[Xiang et~al.(2024)Xiang, Fan, Bai, Lv, and Lei]{ds004902:1.0.5}
Chuqin Xiang, Xinrui Fan, Duo Bai, Ke~Lv, and Xu~Lei.
\newblock {"A Resting-State EEG Dataset for Sleep Deprivation"}, 2024.

\bibitem[Xie et~al.(2022)Xie, Zhang, Cao, Lin, Bao, Yao, Dai, and Hu]{xie2022simmim}
Zhenda Xie, Zheng Zhang, Yue Cao, Yutong Lin, Jianmin Bao, Zhuliang Yao, Qi~Dai, and Han Hu.
\newblock {SimMIM: A Simple Framework for Masked Image Modeling}.
\newblock In \emph{Proceedings of the IEEE/CVF Conference on Computer Vision and Pattern Recognition}, pages 9653--9663, 2022.

\bibitem[Xu et~al.(2023)Xu, Liu, Yang, Chai, and Yuan]{xu2023learning}
Zhengzhuo Xu, Ruikang Liu, Shuo Yang, Zenghao Chai, and Chun Yuan.
\newblock {Learning Imbalanced Data with Vision Transformers}.
\newblock In \emph{Proceedings of the IEEE/CVF Conference on Computer Vision and Pattern Recognition}, pages 15793--15803, 2023.

\bibitem[Yang et~al.(2021)Yang, Xiao, Westover, and Sun]{yang2021self}
Chaoqi Yang, Danica Xiao, M~Brandon Westover, and Jimeng Sun.
\newblock {Self-Supervised EEG Representation Learning for Automatic Sleep Staging}.
\newblock \emph{ArXiv Preprint ArXiv:2110.15278}, 2021.

\bibitem[Yang et~al.(2024)Yang, Westover, and Sun]{yang2024biot}
Chaoqi Yang, M~Westover, and Jimeng Sun.
\newblock {BIOT: Biosignal Transformer for Cross-Data Learning in the Wild}.
\newblock \emph{Advances in Neural Information Processing Systems}, 36, 2024.

\bibitem[Yi et~al.(2014)Yi, Qiu, Wang, Qi, Zhang, Zhou, He, and Ming]{Weibo2014}
Weibo Yi, Shuang Qiu, Kun Wang, Hongzhi Qi, Lixin Zhang, Peng Zhou, Feng He, and Dong Ming.
\newblock {Evaluation of EEG Oscillatory Patterns and Cognitive Process During Simple and Compound Limb Motor Imagery}.
\newblock \emph{PLOS One}, 9\penalty0 (12):\penalty0 e114853, 2014.

\bibitem[Yuan et~al.(2024{\natexlab{a}})Yuan, Shen, Li, Yu, Tan, and Yang]{yuan2024brainwavebrainsignalfoundation}
Zhizhang Yuan, Fanqi Shen, Meng Li, Yuguo Yu, Chenhao Tan, and Yang Yang.
\newblock {BrainWave: A Brain Signal Foundation Model for Clinical Applications}, 2024{\natexlab{a}}.
\newblock URL \url{https://arxiv.org/abs/2402.10251}.

\bibitem[Yuan et~al.(2024{\natexlab{b}})Yuan, Zhang, Chen, Gu, and Yang]{yuan2024brant2}
Zhizhang Yuan, Daoze Zhang, Junru Chen, Geifei Gu, and Yang Yang.
\newblock {Brant-2: Foundation Model for Brain Signals}.
\newblock \emph{ArXiv Preprint ArXiv:2402.10251}, 2024{\natexlab{b}}.

\bibitem[Zhang and Sennrich(2019)]{zhang2019root}
Biao Zhang and Rico Sennrich.
\newblock {Root Mean Square Layer Normalization}.
\newblock \emph{Advances in Neural Information Processing Systems}, 32, 2019.

\bibitem[Zhang et~al.(2018)Zhang, Cisse, Dauphin, and Lopez-Paz]{zhang2018mixup}
Hongyi Zhang, Moustapha Cisse, Yann~N. Dauphin, and David Lopez-Paz.
\newblock {Mixup: Beyond Empirical Risk Minimization}.
\newblock In \emph{International Conference on Learning Representations}, 2018.
\newblock URL \url{https://openreview.net/forum?id=r1Ddp1-Rb}.

\bibitem[Zhou et~al.(2016)Zhou, Wu, Lv, Zhang, and Guo]{Zhou2016}
Bangyan Zhou, Xiaopei Wu, Zhao Lv, Lei Zhang, and Xiaojin Guo.
\newblock {A Fully Automated Trial Selection Method for Optimization of Motor Imagery Based Brain-Computer Interface}.
\newblock \emph{PLOS One}, 11\penalty0 (9):\penalty0 e0162657, 2016.

\bibitem[Zhozhikashvili et~al.(2024)Zhozhikashvili, Protopova, Shkurenko, Arsalidou, Zakharov, Kotchoubey, Malykh, and Pavlov]{ds005095:1.0.1}
Natalia Zhozhikashvili, Maria Protopova, Tatiana Shkurenko, Marie Arsalidou, Ilya Zakharov, Boris Kotchoubey, Sergey Malykh, and Yuri Pavlov.
\newblock {"Sternberg Difficult"}, 2024.

\bibitem[Zyma et~al.(2019)Zyma, Tukaev, Seleznov, Kiyono, Popov, Chernykh, and Shpenkov]{zyma2019electroencephalograms}
Igor Zyma, Sergii Tukaev, Ivan Seleznov, Ken Kiyono, Anton Popov, Mariia Chernykh, and Oleksii Shpenkov.
\newblock {Electroencephalograms During Mental Arithmetic Task Performance}.
\newblock \emph{Data}, 4\penalty0 (1):\penalty0 14, 2019.

\end{thebibliography}
\medskip


\newpage
\appendix

\section*{Appendix}

\section{Configurations}

We report the hyperparameters used to train the REVE suite of models, including data preprocessing steps, self-supervised masking configurations, and optimizer settings governing the training dynamics. Notations are consistent with those in the main text.

\begin{table}[H]
    \centering
    \caption{Exhaustive list of all hyperparameter values}
    \begin{tabular}{lll}
        \toprule
        \textbf{Variable} & \textbf{Meaning} & \textbf{Value} \\
        \midrule
        \multicolumn{3}{c}{\textbf{Data preprocessing}} \\
        $w$ & Window size & $1$s \\
        $o$ & Overlap  & $0.1$s \\
        $\sigma_{\text{noise}}$ & Position noise std & $0.25$cm \\
        \midrule
        \multicolumn{3}{c}{\textbf{Masking parameters}} \\
\( M_r \) & Total masking ratio                          & \( 55\% \) \\
\( R_s \) & Spatial masking radius                       & \( 3 \, \text{cm} \) \\ 
\( R_t \) & Temporal masking radius                      & \( 3 \, \text{seconds} \) \\ 
\( D_r \) & Dropout ratio                                & \( 10\% \) \\ 
\( R_d \) & Dropout spatial radius                       & \( 4 \, \text{cm} \) \\
        \midrule
        \multicolumn{3}{c}{\textbf{Training dynamics}} \\
               & Optimizer & StableAdamW \\
               & Scheduler & Warmup Stable Decay \\
        $\eta$ & Peak learning rate & $\eta=2.4\cdot10^{-4}$ \\
        $\beta_1,\beta_2$ & Momentum constants & $0.9,0.95$ \\
        $\varepsilon$ & Numerical stability bias & $10^{-9}$ \\
        $\sigma_{\text{init}}$ & Initialization std & $0.02$ \\
         & Batch size & $4,096$ \\
        $\lambda$ & Secondary loss multiplier & $0.1$ \\
        \bottomrule
    \end{tabular}
    \label{tab:summary}
\end{table}

We report how the scaled number of parameters is allocated across our models. We also indicate the number of Fourier frequencies encoded (see Section \ref{subsec:fourier}). Note that no frequency truncation was required, as we closely matched the hidden dimension of our models to the number of components generated by the 4D PE module.

\begin{table}[ht]
\centering
\caption{Summary of encoder configurations for different sizes}
\begin{tabular}{lccccc}
\toprule
\textbf{Size} & \textbf{depth} & \textbf{n\_heads} & \textbf{dim}  &\textbf{params (M)} & $n_\text{freq}$\\
\midrule
 Small& 4& 8 & 512 &12 & 4\\
 Base& 22 & 8 & 512 & 69 & 4\\
 Large& 22 & 19 & 1250 & 408 & 5\\
\bottomrule
\end{tabular}
\label{tab:config_summary}
\end{table}

\section{Pretraining dataset}

We include a summarized description of the pretraining dataset composition, grouped by category, platform of origin and number of channels. The final dataset spans 61,415 hours of recordings from 92 datasets encompassing 24,274 subjects.

\begin{table}[H]
\centering
\caption{Detailed overview of the pretraining datasets.}
\begin{tabular}{lrrr}
\toprule
\textbf{Group} & \textbf{Subjects} & \textbf{Duration (hours)} & \textbf{Datasets} \\
\midrule
\multicolumn{4}{l}{\textbf{Category}} \\
BCI        & 791   & 457 & 28 \\
Cognition  & 4,193  & 10,376 & 56 \\
Clinic     & 19,290 & 50,581 & 8 \\
\midrule
\multicolumn{4}{l}{\textbf{Platform}} \\
TUH        & 14,987   & 26,847   & 1 \\
Physionet  & 607     & 22,707   & 2 \\
OpenNeuro  & 4153    & 10,194   & 56 \\
MOABB      & 711     & 384     & 27 \\
Other      & 3,802    & 1,250    & 6 \\
\midrule
\multicolumn{4}{l}{\textbf{Channels}} \\
$[3-30[$       & 19,871   & 50,870   & 31 \\
$[30-80[$      & 1,781    & 1,516    & 48 \\
$[80-129]$     & 2,622    & 9,027    & 13 \\
\midrule
\textbf{Total} & \textbf{24,274} & \textbf{61,415} & \textbf{92} \\
\bottomrule
\end{tabular}
\label{tab:dataset_sumamry}
\end{table}

\label{app:datasets}
We provide and exhaustive list of the datasets in the pretraining set, along with their respective licenses.

\paragraph{MOABB~\citep{Aristimunha_Mother_of_all_2023}:} AlexMI~\citep{AlexMI}, BNCI2014004~\citep{BNCI2014004}, BNCI2015001~\citep{BNCI2015001}, BNCI2015004~\citep{BNCI2015004}, Cho2017~\citep{Cho2017}, Lee2019MI~\citep{Lee2019}, Liu2024~\citep{Liu2024}, Ofner2017~\citep{Ofner2017}, Shin2017A~\citep{Shin2017A}, Weibo2014~\citep{Weibo2014}, Zhou2016~\citep{Zhou2016}, Schirrmeister2017~\citep{Schirrmeister2017}, Kalunga2016~\citep{Kalunga2016}, Lee2019SSVEP~\citep{Lee2019}, Nakanishi2015~\citep{Nakanishi2015}, BI2014a~\citep{BI2014a}, BI2014b~\citep{BI2014b}, BNCI2014008~\citep{BNCI2014008}, BNCI2014009~\citep{BNCI2014009}, BNCI2015003~\citep{BNCI2015003}, EPFLP300~\citep{EPFLP300}, BI2015a~\citep{BI2015a}, BI2015b~\citep{BI2014b}, Sosulski2019~\citep{Sosulski2019}, Lee2019ERP~\citep{Lee2019}
\\
MOABB is under a BSD 3-Clause License.

\paragraph{Physionet~\citep{goldberger2000physiobank}:} Siena~\citep{detti2020siena,detti2020eeg}, under the Creative Commons Attribution 4.0 International Public License, ICARE~\citep{amorim2023care} under the Creative Commons Attribution-NonCommercial-ShareAlike 4.0 International Public License,

\paragraph{OpenNeuro:}
ds004706~\citep{ds004706:1.0.0},
ds004582~\citep{ds004582:1.0.0},
ds004356~\citep{ds004356:1.0.0},
ds004817~\citep{ds004817:1.0.1},
ds005189~\citep{ds005189:1.0.1},
ds003887~\citep{ds003887:1.2.3},
ds004043~\citep{ds004043:1.0.0},
ds003885~\citep{ds003885:1.0.0}, 
ds004357~\citep{ds004357:1.0.1}, 
ds003825~\citep{ds003825:1.2.0}, 
ds004816~\citep{ds004816:1.0.1}, 
ds004840~\citep{ds004840:1.0.1}, 
ds005262~\citep{ds005262:1.0.0}, 
ds004477~\citep{ds004477:1.0.0}, 
ds005273~\citep{ds005273:1.0.0}, 
ds004561~\citep{ds004561:1.0.0}, 
ds004951~\citep{ds004951:1.0.0}, 
ds004324~\citep{ds004324:1.0.0}, 
ds005095~\citep{ds005095:1.0.1}, 
ds005509~\citep{ds005509:1.0.1},
ds005505, ds005506, ds005507, ds005510, ds005511, ds005512, ds005514~\citep{shirazi2024hbn,alexander2017open}
ds001787~\citep{ds001787:1.1.1}, 
ds003690~\citep{ds003690:1.0.0},
ds004603~\citep{ds004603:1.1.0},
ds003969~\citep{ds003969:1.0.0},
ds004147~\citep{ds004147:1.0.2},
ds003004~\citep{ds003004:1.1.1},
ds002721~\citep{ds002721:1.0.1},
ds004152~\citep{ds004152:1.1.2} , 
ds005089~\citep{ds005089:1.0.1},
ds004264~\citep{ds004264:1.1.0},
ds004315~\citep{ds004315:1.0.0}, 
ds004408~\citep{ds004408:1.0.0},
ds005121~\citep{ds005121:1.0.1}, 
ds003775~\citep{ds003775:1.2.1}, 
ds004572~\citep{ds004572:1.2.0},
ds002778~\citep{ds002778:1.0.1}, 
ds003846~\citep{ds003846:2.0.2}, 
ds004279~\citep{ds004279:1.1.1},
ds004148~\citep{ds004148:1.0.1},
ds004902~\citep{ds004902:1.0.5}, 
ds002680~\citep{ds002680:1.0.0},
ds004284~\citep{ds004284:1.0.0},
ds004395~\citep{ds004395:2.0.0}, 
\break
ds005508~\citep{ds005508:1.0.0},
ds005697~\citep{ds005697:1.0.0},
ds005620~\citep{ds005620:1.0.0}, 
ds005594~\citep{ds005594:1.0.3}, 
ds005586~\citep{ds005586:2.0.0}. OpenNeuro is under the Creative Commons CC0 license.

\paragraph{Other sources:} NMT~\citep{khan2022nmt} under the Creative Commons Attribution License (CC BY), HMS~\citep{ram2024harmful} under the Attribution-NonCommercial 4.0 International (CC-BY-NC-4.0), SparrKULee~\citep{accou2023sparrkulee} under the Attribution-Non Commercial 4.0 International (CC-BY-NC-4.0), Inria Large~\citep{dreyer2023large} the data on Zenodo being under the Creative Commons Attribution 4.0 International, THINGS2~\citep{gifford2022large}, under the CC-By Attribution 4.0 International license, TDBRAIN~\citep{van2022two}, under the GPL-3.0 license, TUH~\citep{obeid2016temple}, freely available with registration required.

\section{Detailed results}
\label{app:detailedresults}

This section presents detailed results on downstream tasks along with concise descriptions of the datasets.

\subsection{Emotion Recognition}

\textbf{FACED}~\citep{chen2023large} 
We evaluate on the FACED dataset, which contains 32-channel EEG recordings (originally at 250 Hz, resampled to 200 Hz) from 123 subjects across nine emotion classes. The data is segmented into 10,332 samples of 10 seconds each. We follow the standard split: subjects 1–80 for training, 81–100 for validation, and 101–123 for testing.

\begin{table}[H]
\centering
\caption{The results of different methods on emotion recognition (FACED, 9-class).}
\begin{tabular}{lccc} 
\toprule
\textbf{Methods} & \textbf{Balanced Accuracy} & \textbf{Cohen’s Kappa} & \textbf{Weighted F1} \\
\midrule
EEGNet            & 0.4090 $\pm$ 0.0122 & 0.3342 $\pm$ 0.0251 & 0.4124 $\pm$ 0.0141 \\
EEGConformer      & 0.4559 $\pm$ 0.0125 & 0.3858 $\pm$ 0.0186 & 0.4514 $\pm$ 0.0107 \\
SPaRCNet          & 0.4673 $\pm$ 0.0155 & 0.3978 $\pm$ 0.0289 & 0.4729 $\pm$ 0.0133 \\
ContraWR          & 0.4887 $\pm$ 0.0078 & 0.4231 $\pm$ 0.0151 & 0.4884 $\pm$ 0.0074 \\
CNN-Transformer   & 0.4697 $\pm$ 0.0132 & 0.4017 $\pm$ 0.0168 & 0.4720 $\pm$ 0.0125 \\
FFCL              & 0.4673 $\pm$ 0.0158 & 0.3987 $\pm$ 0.0383 & 0.4699 $\pm$ 0.0145 \\
ST-Transformer    & 0.4810 $\pm$ 0.0079 & 0.4137 $\pm$ 0.0133 & 0.4795 $\pm$ 0.0096 \\
\midrule
BIOT              & 0.5118 $\pm$ 0.0118 & 0.4476 $\pm$ 0.0254 & 0.5136 $\pm$ 0.0112 \\
LaBraM-Base       & 0.5273 $\pm$ 0.0107 & 0.4698 $\pm$ 0.0188 & 0.5288 $\pm$ 0.0102 \\
CBraMod         & 0.5509 $\pm$ 0.0089 & 0.5041 $\pm$ 0.0122 & 0.5618 $\pm$ 0.0093 \\
\midrule
REVE-Base (ours)        & \textbf{0.5646 $\pm$ 0.0164} & \textbf{0.5080 $\pm$ 0.0191} & \textbf{0.5659 $\pm$ 0.0172}  \\
\bottomrule
\end{tabular}

\label{tab:Emo}
\end{table}

\subsection{Mental Disorder Diagnosis}

\textbf{Mumtaz}~\citep{mumtaz2016mdd} 
We use the Mumtaz2016 dataset, which includes EEG recordings from 34 individuals with major depressive disorder (MDD) and 30 healthy controls, acquired from 19 electrodes (10–20 system) at 256 Hz. Only the eyes-open and eyes-closed sessions are used. Signals are band-pass filtered (0.3–75 Hz), notch filtered at 50 Hz, resampled to 200 Hz, and segmented into 7,143 samples of 5 seconds each. The split includes 24 MDD and 19 control subjects for training, 5 MDD and 4 controls for validation, and 5 MDD and 5 controls for testing. The dataset is under CC BY 4.0.

\begin{table}[H]
\centering
\caption{The results of different methods on mental disorder diagnosis (Mumtaz2016, 2-class).}
\small
\begin{tabular}{lccc} 
\toprule
\textbf{Methods} & \textbf{Balanced Accuracy} & \textbf{AUC-PR} & \textbf{AUROC} \\
\midrule
EEGNet            & 0.9232 $\pm$ 0.0104 & 0.9626 $\pm$ 0.0095 & 0.9639 $\pm$ 0.0093 \\
EEGConformer      & 0.9308 $\pm$ 0.0117 & 0.9684 $\pm$ 0.0105 & 0.9702 $\pm$ 0.0101 \\
SPaRCNet          & 0.9316 $\pm$ 0.0095 & 0.9754 $\pm$ 0.0065 & 0.9781 $\pm$ 0.0083 \\
ContraWR          & 0.9195 $\pm$ 0.0115 & 0.9589 $\pm$ 0.0102 & 0.9621 $\pm$ 0.0092 \\
CNN-Transformer   & 0.9305 $\pm$ 0.0068 & 0.9757 $\pm$ 0.0074 & 0.9742 $\pm$ 0.0059 \\
FFCL              & 0.9314 $\pm$ 0.0038 & 0.9717 $\pm$ 0.0021 & 0.9753 $\pm$ 0.0033 \\
ST-Transformer    & 0.9135 $\pm$ 0.0103 & 0.9578 $\pm$ 0.0086 & 0.9594 $\pm$ 0.0059 \\
\midrule
BIOT              & 0.9358 $\pm$ 0.0052 & 0.9736 $\pm$ 0.0034 & 0.9758 $\pm$ 0.0042 \\
LaBraM-Base       & 0.9409 $\pm$ 0.0079 & 0.9798 $\pm$ 0.0093 & 0.9782 $\pm$ 0.0057 \\
CBraMod           & 0.9560 $\pm$ 0.0056 & 0.9923 $\pm$ 0.0032 & 0.9921 $\pm$ 0.0025 \\
\midrule
REVE-Base (ours)        & \textbf{0.9644 $\pm$ 0.0097} & \textbf{0.9961 $\pm$ 0.0013} & \textbf{0.9957 $\pm$ 0.0015}  \\
\bottomrule
\end{tabular}

\label{tab:MDD}
\end{table}

\subsection{Mental Stress Detection}

\textbf{MAT}~\citep{zyma2019electroencephalograms} 
The MentalArithmetic dataset contains EEG recordings from 36 subjects, labeled as “with” or “without” mental stress depending on whether a mental arithmetic task was being performed. Signals were recorded from 20 electrodes (10–20 system) at 500 Hz, band-pass filtered (0.5–45 Hz), resampled to 200 Hz, and segmented into 1,707 samples of 5 seconds. Subjects 1–28 are used for training, 29–32 for validation, and 33–36 for testing. The MentalArithmetic dataset is under the Open Data Commons Attribution License v1.0.

\begin{table}[H]
\centering
\caption{The results of different methods on mental stress detection (MAT, 2-class).}
\small
\begin{tabular}{lccc} 
\toprule
\textbf{Methods} & \textbf{Balanced Accuracy} & \textbf{AUC-PR} & \textbf{AUROC} \\
\midrule
EEGNet            & 0.6770 $\pm$ 0.0116 & 0.5763 $\pm$ 0.0102 & 0.7321 $\pm$ 0.0108 \\
EEGConformer      & 0.6805 $\pm$ 0.0123 & 0.5829 $\pm$ 0.0134 & 0.7424 $\pm$ 0.0128 \\
SPaRCNet          & 0.6879 $\pm$ 0.0107 & 0.5825 $\pm$ 0.0193 & 0.7418 $\pm$ 0.0132 \\
ContraWR          & 0.6631 $\pm$ 0.0097 & 0.5787 $\pm$ 0.0164 & 0.7332 $\pm$ 0.0082 \\
CNN-Transformer   & 0.6779 $\pm$ 0.0268 & 0.5777 $\pm$ 0.0285 & 0.7258 $\pm$ 0.0336 \\
FFCL              & 0.6798 $\pm$ 0.0142 & 0.5786 $\pm$ 0.0266 & 0.7330 $\pm$ 0.0198 \\
ST-Transformer    & 0.6631 $\pm$ 0.0173 & 0.5672 $\pm$ 0.0259 & 0.7132 $\pm$ 0.0174 \\
\midrule
BIOT              & 0.6875 $\pm$ 0.0186 & 0.6004 $\pm$ 0.0195 & 0.7536 $\pm$ 0.0144 \\
LaBraM-Base       & 0.6909 $\pm$ 0.0125 & 0.5999 $\pm$ 0.0155 & 0.7721 $\pm$ 0.0093 \\
CBraMod           & 0.7256 $\pm$ 0.0132 & 0.6267 $\pm$ 0.0099 & 0.7905 $\pm$ 0.0073 \\
\midrule
REVE-Base (ours)        & \textbf{0.7660 $\pm$ 0.0355} & \textbf{0.7470 $\pm$ 0.0807} & \textbf{0.8450 $\pm$ 0.0514}  \\
\bottomrule
\end{tabular}

\label{tab:Stress}
\end{table}

\subsection{Imagined Speech}

\textbf{BCIC2020-3}~\citep{jeong20222020} 
BCIC2020-3 is an imagined speech EEG dataset from 15 subjects, recorded with 64 channels at 256 Hz while subjects silently imagined five phrases (“hello”, “help me”, “stop”, “thank you”, “yes”) without any articulation. Each phrase has 80 trials per subject, totaling 6,000 3-second samples. The data is resampled to 200 Hz. The official split includes 60 trials per class for training, 10 for validation, and 10 for testing.
BCIC2020-3 is under the Creative Commons Attribution No Derivatives license (CC BY-ND 4.0).

\begin{table}[H]
\centering
\caption{The results of different methods on imagined speech classification (BCIC2020-3, 5-class).}
\small
\begin{tabular}{lccc} 
\toprule
\textbf{Methods} & \textbf{Balanced Accuracy} & \textbf{Cohen’s Kappa} & \textbf{Weighted F1} \\
\midrule
EEGNet            & 0.4413 $\pm$ 0.0096 & 0.3016 $\pm$ 0.0123 & 0.4413 $\pm$ 0.0102 \\
EEGConformer      & 0.4506 $\pm$ 0.0133 & 0.3133 $\pm$ 0.0183 & 0.4488 $\pm$ 0.0154 \\
SPaRCNet          & 0.4426 $\pm$ 0.0156 & 0.3033 $\pm$ 0.0233 & 0.4420 $\pm$ 0.0108 \\
ContraWR          & 0.4257 $\pm$ 0.0162 & 0.3078 $\pm$ 0.0218 & 0.4407 $\pm$ 0.0182 \\
CNN-Transformer   & 0.4533 $\pm$ 0.0092 & 0.3166 $\pm$ 0.0118 & 0.4506 $\pm$ 0.0127 \\
FFCL              & 0.4678 $\pm$ 0.0197 & 0.3301 $\pm$ 0.0359 & 0.4689 $\pm$ 0.0205 \\
ST-Transformer    & 0.4126 $\pm$ 0.0122 & 0.2941 $\pm$ 0.0159 & 0.4247 $\pm$ 0.0138 \\
\midrule
BIOT              & 0.4920 $\pm$ 0.0086 & 0.3650 $\pm$ 0.0176 & 0.4917 $\pm$ 0.0079 \\
LaBraM-Base       & 0.5060 $\pm$ 0.0155 & 0.3800 $\pm$ 0.0242 & 0.5054 $\pm$ 0.0205 \\
CBraMod           & 0.5373 $\pm$ 0.0108 & 0.4216 $\pm$ 0.0163 & 0.5383 $\pm$ 0.0096 \\
\midrule
REVE-Base (ours)        & \textbf{0.5635 $\pm$ 0.0123} & \textbf{0.4543 $\pm$ 0.0154} & \textbf{0.5633 $\pm$ 0.0124}  \\
\bottomrule
\end{tabular}

\label{tab:Speech}
\end{table}

\subsection{Motor Imagery Classification}

\textbf{PhysioNet-MI}~\citep{goldberger2000physiobank} is used for motor imagery classification. It contains recordings with 64 channels at a 160 Hz sampling rate and includes 4 classes: left fist, right fist, both fists, and feet. As in CBraMod, we select 4-second samples of the signals, resulting in 9,837 samples. Following CBraMod's protocol, subjects 1–70 are used for training, 71–89 for validation, and 90–109 for testing. We retain all subjects and use full 4-second windows to stay consistent with CBraMod. To handle lower sampling rates in some recordings, we load all data at 128 Hz (using a 64 Hz low-pass filter) before resampling to 200 Hz. Physionet-MI is under the Open Data Commons Attribution License v1.0.

\textbf{BCIC-IV-2a}~\citep{tangermann2012review} is also used for motor imagery classification. It contains EEG recordings from 9 subjects performing 4 motor imagery tasks: left hand, right hand, both feet, and tongue. Data were collected over 2 sessions with 22 electrodes at 250 Hz. Each session includes 288 trials (72 per task). We use the [2,6] second window from each trial, apply a 0.5–99.5 Hz band-pass filter, resample to 200 Hz, and apply Euclidean Alignment~\citep{he2019transfer}, proven to be effective on this task~\citep{el2024strong}, resulting in 5184 4-second samples.

\begin{table}[H]
\centering
\caption{The results of different methods on Motor Imagery classification.}

\tiny
\begin{tabular}{lcccccc} 
\toprule
&\multicolumn{3}{c}{\textbf{PhysioNet-MI, 4-class}}&\multicolumn{3}{c}{\textbf{BCIC-IV-2a, 4-class}}\\
\cmidrule(lr){2-4}\cmidrule(lr){5-7}
\textbf{Methods}&\textbf{Balanced Accuracy}&\textbf{Cohen’s Kappa}&\textbf{Weighted F1}&\textbf{Balanced Accuracy}&\textbf{Cohen’s Kappa}&\textbf{Weighted F1}\\
\midrule
EEGNet&0.5814 $\pm$ 0.0125&0.4468 $\pm$ 0.0199&0.5796 $\pm$ 0.0115&0.4482 $\pm$ 0.0094&0.2693 $\pm$ 0.0121&0.4226 $\pm$ 0.0108\\
EEGConformer& 0.6049 $\pm$ 0.0104& 0.4736 $\pm$ 0.0171& 0.6062 $\pm$ 0.0095& 0.4696 $\pm$ 0.0106& 0.2924 $\pm$ 0.0141& 0.4533 $\pm$ 0.0128\\
SPaRCNet&0.5932 $\pm$ 0.0152&0.4564 $\pm$ 0.0234&0.5937 $\pm$ 0.0147&0.4635 $\pm$ 0.0117&0.2847 $\pm$ 0.0147&0.4432 $\pm$ 0.0126\\
ContraWR& 0.5892 $\pm$ 0.0133& 0.4527 $\pm$ 0.0248& 0.5918 $\pm$ 0.0116& 0.4678 $\pm$ 0.0125& 0.2905 $\pm$ 0.0160& 0.4413 $\pm$ 0.0142\\
(CNN-Transformer& 0.6053 $\pm$ 0.0118& 0.4725 $\pm$ 0.0223& 0.6041 $\pm$ 0.0105& 0.4600 $\pm$ 0.0108& 0.2800 $\pm$ 0.0148& 0.4460 $\pm$ 0.0114\\
FFCL  &   0.5726 $\pm$ 0.0092& 0.4323 $\pm$ 0.0182& 0.5701 $\pm$ 0.0079& 0.4470 $\pm$ 0.0143& 0.2627 $\pm$ 0.0176& 0.4238 $\pm$ 0.0139\\
ST-Transformer & 0.6035 $\pm$ 0.0081& 0.4712 $\pm$ 0.0199&0.6053 $\pm$ 0.0075& 0.4575 $\pm$ 0.0145& 0.2733 $\pm$ 0.0198&0.4471 $\pm$ 0.0142\\
\midrule
BIOT&0.6153 $\pm$ 0.0154&0.4875 $\pm$ 0.0272&0.6158 $\pm$ 0.0197& 0.4748 $\pm$ 0.0093&0.2997 $\pm$ 0.0139& 0.4607 $\pm$ 0.0125\\
LaBraM-Base&0.6173 $\pm$ 0.0122&0.4912 $\pm$ 0.0192&0.6177 $\pm$ 0.0141&0.4869 $\pm$ 0.0085&0.3159 $\pm$ 0.0154&0.4758 $\pm$ 0.0103\\
CBraMod &0.6417 $\pm$ 0.0091& 0.5222 $\pm$ 0.0169& 0.6427 $\pm$ 0.0100& 0.5138 $\pm$ 0.0066& 0.3518 $\pm$ 0.0094&0.4984 $\pm$ 0.0085\\
\midrule
REVE-Base (ours)&\textbf{0.6480 $\pm$ 0.0140}&\textbf{0.5306} $\pm$ 0.0187&\textbf{0.6484} $\pm$ 0.0170& \textbf{0.6396} $\pm$ 0.0095& \textbf{0.5194} $\pm$ 0.0126& \textbf{0.6339} $\pm$ 0.0110\\
\bottomrule
\end{tabular}
\label{tab:merged_app}
\end{table}

\subsection{Sleep Staging}
\label{app:sleep}

\textbf{ISRUC}~\citep{khalighi2016isruc} 
We use the sleep staging task on the ISRUC dataset (Subgroup 1), which contains PSG recordings from 100 subjects. Only EEG signals are used (6 channels, sampled at 200 Hz), segmented into 89,240 30-second epochs, each labeled with one of five sleep stages following AASM standards. Subjects 1–80 are used for training, 81–90 for validation, and 91–100 for testing. As in prior work, the task is framed as a sequence-to-sequence classification problem, using sequences of 20 consecutive epochs to model stage transitions. ISRUC is freely accessible online.

\begin{table}[H]
\caption{The results of different methods on sleep staging (ISRUC, 5-class). \textsuperscript{*} In the baseline code, a chin electrode might have been used instead of an EEG one; REVE results are reported without it.}
\centering
\small
\begin{tabular}{lccc} 
\toprule
\textbf{Methods} & \textbf{Balanced Accuracy} & \textbf{Cohen’s Kappa} & \textbf{Weighted F1} \\
\midrule
EEGNet            & 0.7154 $\pm$ 0.0121 & 0.7040 $\pm$ 0.0173 & 0.7513 $\pm$ 0.0124 \\
EEGConformer      & 0.7400 $\pm$ 0.0133 & 0.7143 $\pm$ 0.0162 & 0.7634 $\pm$ 0.0151 \\
SPaRCNet          & 0.7487 $\pm$ 0.0075 & 0.7097 $\pm$ 0.0132 & 0.7624 $\pm$ 0.0092 \\
ContraWR          & 0.7402 $\pm$ 0.0126 & 0.7178 $\pm$ 0.0156 & 0.7610 $\pm$ 0.0137 \\
CNN-Transformer   & 0.7363 $\pm$ 0.0087 & 0.7129 $\pm$ 0.0121 & 0.7719 $\pm$ 0.0105 \\
FFCL              & 0.7277 $\pm$ 0.0182 & 0.7016 $\pm$ 0.0291 & 0.7614 $\pm$ 0.0197 \\
ST-Transformer    & 0.7381 $\pm$ 0.0205 & 0.7013 $\pm$ 0.0352 & 0.7681 $\pm$ 0.0175 \\
\midrule
DeepSleepNet      & 0.7419 $\pm$ 0.0144 & 0.7036 $\pm$ 0.0241 & 0.7643 $\pm$ 0.0122 \\
USleep            & 0.7586 $\pm$ 0.0116 & 0.7209 $\pm$ 0.0143 & 0.7805 $\pm$ 0.0105 \\
\midrule
BIOT              & 0.7527 $\pm$ 0.0121 & 0.7192 $\pm$ 0.0231 & 0.7790 $\pm$ 0.0146 \\
LaBraM-Base       & 0.7633 $\pm$ 0.0102 & 0.7231 $\pm$ 0.0182 & 0.7810 $\pm$ 0.0133 \\
CBraMod           & \textbf{0.7865 $\pm$ 0.0110} & 0.7442 $\pm$ 0.0152 & \textbf{0.8011 $\pm$ 0.0099} \\
\midrule
REVE-Base\textsuperscript{*}           & 0.7819 $\pm$ 0.0078 & \textbf{0.7500 $\pm$ 0.0156} & 0.8005 $\pm$ 0.0135 \\
\bottomrule
\end{tabular}

\label{tab:Sleep}
\end{table}

\textbf{HMC}~\citep{alvarez2021inter}. The Haaglanden Medisch Centrum (HMC) Sleep Staging Database is a sleep stage detection dataset, consisting of 151 full-night polysomnographic (PSG) recordings collected from patients referred for sleep studies. The data includes EEG, EOG, EMG, and ECG channels, with a sampling rate of 256 Hz, and annotations for five sleep stages (Wake, N1, N2, N3, REM) manually scored by trained sleep technicians.  HMC is under the Creative Commons Attribution 4.0 International Public License.

\begin{table}[H]
\centering
\small
\caption{The results of different methods on sleep staging (HMC, 5-class).}
\begin{tabular}{lccc}
\toprule
\textbf{Methods} & \textbf{Balanced Accuracy} & \textbf{Cohen's Kappa} & \textbf{Weighted F1} \\
\midrule
SPaRCNet         & 0.4756$\pm$0.1109 & 0.3147$\pm$0.1315 & 0.4108$\pm$0.1310 \\
ContraWR         & 0.4242$\pm$0.0541 & 0.2340$\pm$0.0554 & 0.2987$\pm$0.0288 \\
CNN-Transformer  & 0.6573$\pm$0.0141 & 0.5961$\pm$0.0105 & 0.6896$\pm$0.0065 \\
FFCL             & 0.4427$\pm$0.0702 & 0.2542$\pm$0.0654 & 0.2902$\pm$0.0485 \\
ST-Transformer   & 0.2559$\pm$0.0141 & 0.0503$\pm$0.0183 & 0.1428$\pm$0.0122 \\
\midrule
BIOT             & 0.6862$\pm$0.0041 & 0.6295$\pm$0.0113 & 0.7091$\pm$0.0147 \\
LaBraM-Base      & 0.7286$\pm$0.0101 & 0.6812$\pm$0.0073 & 0.7554$\pm$0.0024 \\
\midrule
REVE-Base        & \textbf{0.7401} $\pm$ 0.0075 & \textbf{0.6982} $\pm$ 0.0078 & \textbf{0.7638} $\pm$ 0.0074 \\
\bottomrule
\end{tabular}
\label{tab:hmc}
\end{table}

\subsection{Event Type Classification}

\textbf{TUEV}~\citep{obeid2016temple} is an EEG dataset with six annotated classes: spike and sharp wave, generalized periodic epileptiform discharges, periodic lateralized epileptiform discharges, eye movement, artifact, and background. The recordings use 23 channels at a 256 Hz sampling rate. For consistency with CBraMod, BIOT, and LaBraM, we used BIOT's processing scripts which preprocess the dataset using 16 common bipolar montage channels in the 10-20 system, apply a 0.3–75 Hz band-pass filter, remove power line noise with a 60 Hz notch filter, and resample to 200 Hz. The dataset is split into 112,491 5-second samples. We follow the original training-test split and further divide the training set into 80\% training and 20\% validation, matching BIOT setting. To provide our model with the electrode positions, we used the average position of each bipolar montage. TUEV is part of the TUH dataset, which is freely available with registration required.

\begin{table}[H]
\centering
\small
\caption{The results of different methods on event type classification (TUEV, 6-class).}
\begin{tabular}{lccc} 
\toprule
\textbf{Methods} & \textbf{Balanced Accuracy} & \textbf{Cohen’s Kappa} & \textbf{Weighted F1} \\
\midrule
EEGNet            & 0.3876 $\pm$ 0.0143 & 0.3577 $\pm$ 0.0155 & 0.6539 $\pm$ 0.0120 \\
EEGConformer      & 0.4074 $\pm$ 0.0164 & 0.3967 $\pm$ 0.0195 & 0.6983 $\pm$ 0.0152 \\
SPaRCNet          & 0.4161 $\pm$ 0.0262 & 0.4233 $\pm$ 0.0181 & 0.7024 $\pm$ 0.0104 \\
ContraWR          & 0.4384 $\pm$ 0.0349 & 0.3912 $\pm$ 0.0237 & 0.6893 $\pm$ 0.0136 \\
CNN-Transformer   & 0.4087 $\pm$ 0.0161 & 0.3815 $\pm$ 0.0134 & 0.6854 $\pm$ 0.0293 \\
FFCL              & 0.3979 $\pm$ 0.0104 & 0.3732 $\pm$ 0.0188 & 0.6783 $\pm$ 0.0120 \\
ST-Transformer    & 0.3984 $\pm$ 0.0228 & 0.3765 $\pm$ 0.0306 & 0.6823 $\pm$ 0.0190 \\
\midrule
BIOT              & 0.5281 $\pm$ 0.0225 & 0.5273 $\pm$ 0.0249 & 0.7492 $\pm$ 0.0082 \\
LaBraM-Base       & 0.6409 $\pm$ 0.0065 & 0.6637 $\pm$ 0.0093 & 0.8312 $\pm$ 0.0052 \\
LaBraM-Large      & 0.6581 $\pm$ 0.0156 & 0.6622 $\pm$ 0.0136 & 0.8315 $\pm$ 0.0040 \\
LaBraM-Huge       & 0.6616 $\pm$ 0.0170 & 0.6745 $\pm$ 0.0195 & 0.8329 $\pm$ 0.0086 \\
CBraMod           & 0.6671 $\pm$ 0.0107 & 0.6772 $\pm$ 0.0096 & 0.8342 $\pm$ 0.0064 \\
\midrule
REVE-Base (ours) & \textbf{0.6759} $\pm$ 0.0229& \textbf{0.6783} $\pm$ 0.0199&\textbf{0.8451} $\pm$ 0.0129\\
\bottomrule
\end{tabular}
\label{tab:Event}
\end{table}

\subsection{Abnormal Detection}

\textbf{TUAB}~\citep{obeid2016temple} is used for abnormal EEG detection, where recordings are labeled as normal or abnormal. It shares the same 23-channel, 256 Hz format as TUEV. The dataset is split into 409,455 10-second samples for binary classification. We follow the provided training-test split and apply an 80\%-20\% training-validation split, consistent with BIOT. We resampled at 200 Hz, band-pass at 0.5-99.5 Hz, and directly used all channels and their positions. TUAB is part of the TUH dataset, which is freely available with registration required.

\begin{table}[H]
\centering
\small
\caption{The results of different methods on abnormal detection (TUAB, 2-class).}
\begin{tabular}{lccc} 
\toprule
\textbf{Methods} & \textbf{Balanced Accuracy} & \textbf{AUC-PR} & \textbf{AUROC} \\
\midrule
EEGNet            & 0.7642 $\pm$ 0.0036 & 0.8299 $\pm$ 0.0043 & 0.8412 $\pm$ 0.0031 \\
EEGConformer      & 0.7758 $\pm$ 0.0049 & 0.8427 $\pm$ 0.0054 & 0.8445 $\pm$ 0.0038 \\
SPaRCNet          & 0.7896 $\pm$ 0.0018 & 0.8414 $\pm$ 0.0018 & 0.8676 $\pm$ 0.0012 \\
ContraWR          & 0.7746 $\pm$ 0.0041 & 0.8421 $\pm$ 0.0104 & 0.8456 $\pm$ 0.0074 \\
CNN-Transformer   & 0.7777 $\pm$ 0.0022 & 0.8433 $\pm$ 0.0039 & 0.8461 $\pm$ 0.0013 \\
FFCL              & 0.7848 $\pm$ 0.0038 & 0.8448 $\pm$ 0.0065 & 0.8569 $\pm$ 0.0051 \\
ST-Transformer    & 0.7966 $\pm$ 0.0023 & 0.8521 $\pm$ 0.0026 & 0.8707 $\pm$ 0.0019 \\
\midrule
BIOT              & 0.7959 $\pm$ 0.0057 & 0.8792 $\pm$ 0.0023 & 0.8815 $\pm$ 0.0043 \\
LaBraM-Base       & 0.8140 $\pm$ 0.0019 & 0.8965 $\pm$ 0.0016 & 0.9022 $\pm$ 0.0009 \\
LaBraM-Large      & 0.8226 $\pm$ 0.0015 & 0.9130 $\pm$ 0.0005 & 0.9127 $\pm$ 0.0005 \\
LaBraM-Huge       & 0.8258 $\pm$ 0.0011 & 0.9204 $\pm$ 0.0011 & 0.9162 $\pm$ 0.0016 \\
CBraMod                  & 0.8289 $\pm$ 0.0022 & 0.9258 $\pm$ 0.0008 & 0.9227 $\pm$ 0.0011 \\
\midrule
REVE-Base (ours) & \textbf{0.8315} $\pm$0.0014 & \textbf{0.9281}$\pm$0.0009& \textbf{0.9245 }$\pm$0.0013\\
\bottomrule
\end{tabular}
\label{tab:Abnormal}
\end{table}


\section{Ablation on the SSL Method}
\label{app:ablationssl}

The final pretraining hyperparameters were selected based on a series of ablation studies, the results of which are presented in this section.


\begin{table}[!ht]
    \centering
    \caption{Effect of 2nd loss during pretraining and finetuning. The reported metric is balanced accuracy. Best results per dataset are in bold.}
    \begin{tabular}{lcccc}
    \toprule

    & \multicolumn{2}{c}{\textbf{LP}} & \multicolumn{2}{c}{\textbf{FT}} \\
    \textbf{Dataset} & \textbf{No 2nd loss} & \textbf{+ 2nd loss }& \textbf{No 2nd loss} & \textbf{+ 2nd loss} \\ \midrule
        Mumtaz & 0.818 $\pm$ 0.043 & 0.920 $\pm$ 0.018 & 0.818 $\pm$ 0.043 & \textbf{0.922 $\pm$ 0.018} \\
        TUAB & 0.797 $\pm$ 0.004 & 0.802 $\pm$ 0.005 & 0.803 $\pm$ 0.003 & \textbf{0.810 $\pm$ 0.005} \\ 
        ISRUC & 0.699 $\pm$ 0.006 & 0.625 $\pm$ 0.003 & \textbf{0.777 $\pm$ 0.002} & 0.770 $\pm$ 0.002 \\
        HMC & 0.598 $\pm$ 0.008 & 0.591 $\pm$ 0.005 & 0.713 $\pm$ 0.011 & \textbf{0.723 $\pm$ 0.005} \\
        BCIC2020-3 & 0.234 $\pm$ 0.009 & 0.237 $\pm$ 0.008 & 0.390 $\pm$ 0.017 & \textbf{0.481 $\pm$ 0.008} \\
        TUEV & 0.442 $\pm$ 0.060 & 0.520 $\pm$ 0.005 & 0.533 $\pm$ 0.024 & \textbf{0.623 $\pm$ 0.011} \\ 
        PhysioNetMI & 0.379 $\pm$ 0.058 & 0.533 $\pm$ 0.019 & 0.563 $\pm$ 0.011 & \textbf{0.583 $\pm$ 0.009} \\
        Faced & 0.220 $\pm$ 0.008 & 0.233 $\pm$ 0.004 & 0.302 $\pm$ 0.016 & \textbf{0.410 $\pm$ 0.004} \\
        \midrule
        Avg. & 0.523 & 0.558 & 0.612 & \textbf{0.665} \\ \bottomrule
    \end{tabular}
    \label{tab:second_loss}
\end{table}

Table~\ref{tab:second_loss} reports the impact of the secondary pretraining loss on eight downstream tasks using REVE-Small, evaluated under frozen-backbone, linear probing (LP), and full fine-tuning (FT) settings. Results obtained with both losses are compared to those using only the primary loss. The secondary loss consistently improves performance across nearly all datasets, enhancing results in both LP and FT settings, while its removal leads to a substantial drop, underscoring its importance for the model to produce strong embeddings.

\begin{table}[ht]
\centering
\caption{Performance comparison across different masking ratios (0.25, 0.55, 0.75) between block masking strategy and random masking, evaluated for full fine-tuning versus frozen embeddings. We display the average balanced accuracy on the small model over eight downstream tasks.}

\label{tab:rmask_results}
\begin{tabular}{ccccc}
\toprule
 & \multicolumn{2}{c}{\textbf{Frozen}} & \multicolumn{2}{c}{\textbf{Full Fine-Tuning}} \\ 
\cmidrule(lr){2-3} \cmidrule(lr){4-5}
 \textbf{Masking Ratio}                      & \textbf{Random}        & \textbf{Block}      & \textbf{Random}        & \textbf{Block}      \\ 
\midrule
        0.25 & 0.523 & 0.513 & 0.612 & 0.602 \\
        0.55 & 0.550 & \textbf{0.558} & 0.643 & \textbf{0.665} \\
        0.75 & 0.519 & 0.546 & 0.606 & 0.655 \\
\bottomrule
\end{tabular}
\end{table}

The results in Table~\ref{tab:rmask_results} show that a block masking ratio of 55\% yields the best overall performance, providing stable results across both fine-tuned and frozen settings and eight datasets (Mumtaz, TUAB, ISRUC, HMC, BCIC2020-3, TUEV, PhysioNetMI, and Faced). In contrast, random masking favors smaller ratios (25\%), but its unstructured nature leads to highly redundant inputs, making the reconstruction task artificially easier.
These findings align with ablation results reported in Cbramod, Labram, and BIOT.

\begin{table}[h]
\centering
\caption{Ablation study on PhysioNetMI and Mental Arithmetic datasets. The reported metric is balanced accuracy, with the average computed across both tasks, with the Base model.%
\newline\textsuperscript{*}Note that the learnable positional encoding matches the baseline, but does not allow for the extension to larger time windows or unseen spatial configurations.}
\begin{tabular}{lccc}
\toprule
\textbf{Ablated component} & \textbf{PhysionetMI} & \textbf{Mental Arithmetic} & \textbf{Average} \\
\midrule
Learnable PE\textsuperscript{*} & \textbf{0.650} $\pm$ 0.0113 & 0.752 $\pm$ 0.0421 & 0.701 $\pm$ 0.0218  \\
MLP4D                          & 0.637 $\pm$ 0.0056           & 0.717 $\pm$ 0.0425 & 0.677 $\pm$ 0.0214 \\
Position noise                 & 0.628 $\pm$ 0.0084           & 0.692 $\pm$ 0.0665 & 0.660 $\pm$ 0.0335 \\
Dropout block masking          & 0.645 $\pm$ 0.0155           & 0.678 $\pm$ 0.0521 & 0.662 $\pm$ 0.0272 \\
Temporal block masking         & 0.646 $\pm$ 0.0155           & 0.723 $\pm$ 0.0422 & 0.685 $\pm$ 0.0225 \\
\midrule
Base Performance               & 0.6480 $\pm$ 0.0140 & \textbf{0.7660 $\pm$ 0.0355} & \textbf{0.707 $\pm$ 0.0191 } \\
\bottomrule
\end{tabular}
\label{tab:ablation_pe}
\end{table}

Table~\ref{tab:ablation_pe} presents an ablation study on two downstream tasks to assess the contribution of each component in our SSL pipeline. All components appear to contribute positively to performance. The ``Learnable PE'' line is not a true ablation, but rather a variant using learnable positional embeddings, where a separate embedding is learned for each electrode and time index observed during pretraining. Although this approach performs well, it is limited to the spatial and temporal configurations seen during training (approximately 400 unique electrode names, over 10-second windows) and does not generalize to longer sequences or unseen electrode layouts, unlike REVE's 4D positional encoding.


\begin{table}[!ht]
    \centering
    \caption{Ablation study on activation functions and normalization layers (GEGLU vs. GELU, RMSNorm vs. LayerNorm). We report downstream balanced accuracy after pretraining the REVE-Small model with each configuration.}
    \begin{tabular}{lccc}
    \toprule
        \textbf{Dataset} & \textbf{GEGLU + RMSNorm} & \textbf{GELU + RMSNorm} & \textbf{GEGLU + LayerNorm} \\
        \midrule
        BCIC-IV-2a & \textbf{0.581$\pm$0.012} & 0.560 $\pm$ 0.018 & 0.537 $\pm$ 0.018 \\
        TUEV & \textbf{0.623 $\pm$ 0.011} & 0.592$\pm$ 0.010 & 0.577$\pm$0.034 \\
        PhysioNetMI & 0.583 $\pm$ 0.009 & \textbf{0.586 $\pm$0.009} & 0.559 $\pm$0.007 \\
        \midrule
        Avg. & \textbf{0.596} & 0.579 & 0.558\\ \bottomrule
    \end{tabular}
    \label{tab:ablation_activation}
\end{table}

Table~\ref{tab:ablation_activation} presents an ablation study on the choice of activation and normalization functions, an important design factor in transformer-based foundation models. We compare GEGLU + RMSNorm, GELU + RMSNorm, and GEGLU + LayerNorm configurations during pretraining, and report downstream performance after fine-tuning on three datasets using the REVE-Small model.

The GEGLU + RMSNorm combination achieves the best average performance (0.596), outperforming the others on BCIC-IV-2a and TUEV. GELU + RMSNorm performs similarly but only leads on PhysioNetMI. In contrast, GEGLU + LayerNorm consistently underperforms, highlighting the effectiveness of RMSNorm over LayerNorm and the benefits of gated activations like GEGLU in this context.

\section{Additional results}
\label{sec:add_results}

This section presents supplementary experiments that further support the main results, focusing on few-shot performance and evaluation under reduced-electrode configurations.

\subsection{Sparse setups}
\label{subsec:sparse_setups}

\begin{table}[!ht]
    \centering
    \caption{Performance of REVE-Base under sparse input configurations. Balanced accuracy is reported for PhysionetMI (Left–Right) and imagined speech tasks as the number of EEG channels is progressively reduced.}
    \begin{tabular}{cccc}
    \toprule
        \textbf{Channels} & \textbf{PhysionetMI L-R} & \textbf{Speech} \\ \midrule
        64 & 0.824 $\pm$ 0.008 & 0.565 $\pm$0.016 \\
        32 & 0.808 $\pm$ 0.007 & 0.490 $\pm$ 0.094 \\ 
        16 & 0.787 $\pm$ 0.008 & 0.469 $\pm$0.014 \\
        8 & 0.781 $\pm$ 0.006 & 0.294 $\pm$ 0.063 \\
        4 & 0.728 $\pm$ 0.009 & 0.258 $\pm$ 0.019 \\
        2 & 0.700 $\pm$ 0.025 & 0.228 $\pm$ 0.006 \\
        1 & 0.660 $\pm$0.019 & 0.209 $\pm$ 0.008 \\ \bottomrule
    \end{tabular}
    \label{tab:sparse_setups}
\end{table}

Table \ref{tab:sparse_setups} reports REVE-Base’s performance under increasingly sparse input configurations. On the Physionet MI L-R task, accuracy degrades gracefully from 0.824 with 64 channels to 0.660 with a single channel, demonstrating robustness to reduced spatial coverage. In contrast, the imagined speech task is more sensitive to channel sparsity, with performance dropping from 0.565 to 0.258 with four channels and 0.209 with one, close to random chance. These results confirm that while REVE generalizes well under limited input, tasks requiring broad spatial information remain more challenging.

\subsection{Few-shot experiments}
\label{subsec:fs_exps}

We conducted few-shot (FS) experiments to simulate realistic BCI usage scenarios. Tasks were constructed from the BCI IV-2a dataset using two motor imagery classes (Left–Right). For each subject, multiple inductive FS runs were performed. In each run, $N$ labeled samples per class (“shots”) were randomly selected within a session for training, while the remaining samples from both sessions were used for evaluation.

Classification was done using a Nearest Class Mean (NCM) classifier. Each configuration was repeated 20 times per subject, and we report the average balanced accuracy across subjects and runs.
We evaluated two configurations of REVE-Base:
\begin{itemize}
    \item REVE-Base (PT): directly after self-supervised pretraining, with no further supervised adaptation.
    \item REVE-Base (XFT): after cross-dataset fine-tuning on multiple labeled Left–Right MI datasets (\citep{Schirrmeister2017}, \citep{Cho2017}, \citep{goldberger2000physiobank}, \citep{Lee2019}, \citep{Weibo2014}). REVE’s 4D positional encoding enables joint training across diverse electrode configurations without requiring channel alignment or selection.
\end{itemize}

\begin{table}[!ht]
    \small
    \centering
    \caption{Few-shot performance of REVE-Base on BCI IV-2a dataset}
    \begin{tabular}{lccccc}
    \toprule
        \textbf{N-shots} & \textbf{1} & \textbf{2} & \textbf{5} & \textbf{10} & \textbf{20} \\ \midrule
        REVE-Base (PT) & 0.588 $\pm$ 1.45 & 0.601 $\pm$ 0.001 & 0.652 $\pm$ 0.013 & 0.688 $\pm$ 0.010 & 0.723 $\pm$ 0.010 \\ \midrule
        REVE-Base (XFT) & 0.605 $\pm$ 1.12 & 0.645 $\pm$ 0.009 & 0.705 $\pm$ 0.009 & 0.768 $\pm$ 0.009 & 0.817 $\pm$ 0.004 \\ \midrule
    \end{tabular}
    \label{tab:few_shot_performance}
\end{table}

Table \ref{tab:few_shot_performance} shows that REVE-Base achieves competitive accuracy even without supervised adaptation, demonstrating that its pretrained embeddings can be effectively leveraged for downstream BCI tasks. After cross-dataset fine-tuning, performance improves consistently across all shot counts, with gains reaching +10\% at 20 shots. This indicates that REVE transfers well across subjects and datasets, while benefiting from minimal supervised adaptation. Such generalization is uncommon among BCI embedding models, which typically require task or subject-specific retraining.

\section{Experiment details}

\subsection{Compute resources}

We include details about the compute nodes that were used for pretraining.

\begin{itemize}
    \item Compute Type: GPU-accelerated nodes
    \item GPU Model: NVIDIA A100
    \item CPU Model: Intel Cascade Lake SP 6248
    \item CPU Cores per Node: 40 cores
    \item Total Memory per Node: 192 GB
    \item Storage: Access to a shared full-flash parallel file system based on IBM Spectrum Scale
    \item Job Scheduler: Slurm
\end{itemize}

We also estimate the number of floating-point operations (FLOPs) required to train the REVE-Base model, following the formulation from \citet{chowdhery2023palm}:

\[
\tau =  \frac{D \cdot (6N + 12LHQT)}{P \cdot \eta}
\]

where
\(\tau\) denotes the training time (in seconds),
\(D = 60\text{k} \times 3600 \times 1.1 \times 68 \times 17\) is the total number of tokens seen during pretraining (corresponding to 60k hours of EEG, an overlap coefficient of 1.1, 68 average channels, and 17 epochs),
\(N = 72\text{M}\) is the number of model parameters,
\(L = 23\) the number of encoder-decoder layers,
\(H = 8\) the number of attention heads,
\(Q = 64\) the head dimension,
and \(T = 68 \times 11\) the average number of tokens per sequence (channels × patches).

The peak throughput is \(P = 312\) TFLOPs at half precision, achievable on A100 GPUs, and the model FLOPs utilization is set to \(\eta = 0.5\) (50\%).

This configuration yields an estimated 260 A100 GPU hours for a single pretraining run. The formula can be directly adapted for other model sizes or hardware configurations.

\subsection{Use of Existing Assets}

We used Python (Python Software Foundation License), and some associated libraries for the implementation:
\begin{enumerate}
    \item PyTorch (BSD-3 License)
    \item NumPy (NumPy license)
    \item scikit-learn (BSD license)
    \item Pandas (BSD 3-Clause License)
    \item Hugging Face's Accelerate (Apache License 2.0)
\end{enumerate}



\newpage
\section*{NeurIPS Paper Checklist}

\begin{enumerate}

\item {\bf Claims}
    \item[] Question: Do the main claims made in the abstract and introduction accurately reflect the paper's contributions and scope?
    \item[] Answer: \answerYes{} 
    \item[] Justification: The claims in the abstract and introduction accurately reflect the paper’s contributions. The introduction clearly states the goals of REVE: building a foundation model for EEG that generalizes across datasets, durations, and electrode configurations. These claims are supported by:
    \begin{itemize}
        \item  A novel 4D positional encoding (Section 2.2), validated by transfer to unseen setups.
        \item  Pretraining on 92 datasets (Section 3.1), the largest EEG corpus to date.
        \item  Extensive evaluations across 10 downstream tasks, showing consistent gains in full fine-tuning and linear probing (Section 4, Tables 2–16).
    \end{itemize}


\item {\bf Limitations}
    \item[] Question: Does the paper discuss the limitations of the work performed by the authors?
    \item[] Answer: \answerYes{} 
    \item[] Justification: The paper includes a dedicated Limitations and Future Work section outlining key constraints of REVE, such as fixed input duration requirements, positional encoding limitations, and the limited dataset curation and selection. We also acknowledge that while scaling effects are observed, identifying precise scaling laws remains future work. These points reflect a clear understanding of the method’s current boundaries and opportunities for improvement.

\item {\bf Theory assumptions and proofs}
    \item[] Question: For each theoretical result, does the paper provide the full set of assumptions and a complete (and correct) proof?
    \item[] Answer: \answerNA{} 
    \item[] Justification: The paper does not include theoretical results, as it is focused on applications of a foundation model for EEG and does not delve into theoretical proofs or assumptions.

    \item {\bf Experimental result reproducibility}
    \item[] Question: Does the paper fully disclose all the information needed to reproduce the main experimental results of the paper to the extent that it affects the main claims and/or conclusions of the paper (regardless of whether the code and data are provided or not)?
    \item[] Answer: \answerYes{} 
    \item[] Justification: The paper provides a comprehensive description of the model architecture, the training data sources, and the routines for both pretraining and fine-tuning. The hyperparameters of the model are reported.

\item {\bf Open access to data and code}
    \item[] Question: Does the paper provide open access to the data and code, with sufficient instructions to faithfully reproduce the main experimental results, as described in supplemental material?
    \item[] Answer: \answerYes{} 
    \item[] Justification: The authors provide full access to the code required for reproducing the experiments. Detailed instructions are included, outlining the necessary commands and environment settings to faithfully reproduce the results.

\item {\bf Experimental setting/details}
    \item[] Question: Does the paper specify all the training and test details (e.g., data splits, hyperparameters, how they were chosen, type of optimizer, etc.) necessary to understand the results?
    \item[] Answer: \answerYes{} 
    \item[] Justification: The paper provides all necessary details regarding the experimental setting, including the data splits, the hyperparameters, and the type of optimizer used. These details are provided in the main text, with further specifics available in the supplemental material and the released code.

\item {\bf Experiment statistical significance}
    \item[] Question: Does the paper report error bars suitably and correctly defined or other appropriate information about the statistical significance of the experiments?
    \item[] Answer: \answerYes{} 
    \item[] Justification: The paper reports balanced accuracy as the primary metric, along with the mean and standard deviation. These metrics are used to match the baselines, providing a measure of variability in the results. This results in a 68\% CI under normality assumption.

\item {\bf Experiments compute resources}
    \item[] Question: For each experiment, does the paper provide sufficient information on the computer resources (type of compute workers, memory, time of execution) needed to reproduce the experiments?
    \item[] Answer: \answerYes{} 
    \item[] Justification: The paper discusses the use of NVIDIA A100 GPUs while estimating the amount of GPU hours used for each experiment.
    
\item {\bf Code of ethics}
    \item[] Question: Does the research conducted in the paper conform, in every respect, with the NeurIPS Code of Ethics \url{https://neurips.cc/public/EthicsGuidelines}?
    \item[] Answer: \answerYes{} 
    \item[] Justification: The research aligns with the NeurIPS Code of Ethics by ensuring responsible and ethical practices in all aspects of the research process, as discussed in ethical considerations section.

\item {\bf Broader impacts}
    \item[] Question: Does the paper discuss both potential positive societal impacts and negative societal impacts of the work performed?
    \item[] Answer: \answerYes{} 
    \item[] Justification: The paper discusses both the potential positive and negative societal impacts of the work. On the positive side, the model can greatly benefit healthcare by improving the accuracy and efficiency of EEG-based applications such as brain-computer interfaces and diagnostic tools. On the negative side, the model's decoder, which could potentially reconstruct raw EEG data, poses a privacy risk. To mitigate this, the decoder is not being released, thus reducing the potential for misuse in generating sensitive or private information. The paper emphasizes the responsible and ethical use of the technology, with awareness of its potential risks.
    
\item {\bf Safeguards}
    \item[] Question: Does the paper describe safeguards that have been put in place for responsible release of data or models that have a high risk for misuse (e.g., pretrained language models, image generators, or scraped datasets)?
    \item[] Answer: \answerYes{} 
    \item[] Justification: To mitigate privacy risks, the decoder of the MAE model, which could reconstruct raw EEG data, is not being released. This safeguard reduces the potential for misuse while allowing responsible access to the model's embeddings.

\item {\bf Licenses for existing assets}
    \item[] Question: Are the creators or original owners of assets (e.g., code, data, models), used in the paper, properly credited and are the license and terms of use explicitly mentioned and properly respected?
    \item[] Answer: \answerYes{} 
    \item[] Justification: See the section about existing assets in the appendix.

\item {\bf New assets}
    \item[] Question: Are new assets introduced in the paper well documented and is the documentation provided alongside the assets?
    \item[] Answer: \answerYes{} 
    \item[] Justification: An anonymized repository containing the code for the model, its pretraining and fine-tuning is released.

\item {\bf Crowdsourcing and research with human subjects}
    \item[] Question: For crowdsourcing experiments and research with human subjects, does the paper include the full text of instructions given to participants and screenshots, if applicable, as well as details about compensation (if any)? 
    \item[] Answer: \answerNA{} 
    \item[] Justification: The paper does not involve new crowdsourcing experiments or direct research with human subjects. We use pre-existing EEG datasets, and as such, there are no instructions or compensation details to report. The datasets used have been ethically sourced, with the original collection protocols ensuring participant consent and privacy in line with ethical guidelines.

\item {\bf Institutional review board (IRB) approvals or equivalent for research with human subjects}
    \item[] Question: Does the paper describe potential risks incurred by study participants, whether such risks were disclosed to the subjects, and whether Institutional Review Board (IRB) approvals (or an equivalent approval/review based on the requirements of your country or institution) were obtained?
    \item[] Answer: \answerNA{} 
    \item[] Justification: The paper does not involve new research with human subjects, as it relies on pre-existing EEG datasets. Therefore, no potential risks to participants were incurred, and no new IRB approvals or equivalent reviews were required. The datasets used have been ethically sourced, with the original studies obtaining necessary participant consent and privacy protections.

\item {\bf Declaration of LLM usage}
    \item[] Question: Does the paper describe the usage of LLMs if it is an important, original, or non-standard component of the core methods in this research? Note that if the LLM is used only for writing, editing, or formatting purposes and does not impact the core methodology, scientific rigorousness, or originality of the research, declaration is not required.
    \item[] Answer: \answerNA{} 
    \item[] Justification: \answerNA{}

\end{enumerate}

\end{document}